# What Can ResNet Learn Efficiently, Going Beyond Kernels?

Zeyuan Allen-Zhu
zeyuan@csail.mit.edu
Microsoft Research AI

Yuanzhi Li
yuanzhil@stanford.edu
Stanford University

May 23, 2019

(version 3)[*]

**Abstract**

How can neural networks such as ResNet *efficiently* learn CIFAR-10 with test accuracy more than 96%, while other methods, especially kernel methods, fall relatively behind? Can we more provide theoretical justifications for this gap?

Recently, there is an influential line of work relating neural networks to kernels in the over-parameterized regime, proving they can learn certain concept class that is also learnable by kernels with similar test error. Yet, can neural networks provably learn some concept class *better* than kernels?

We answer this positively in the distribution-free setting. We prove neural networks can efficiently learn a notable class of functions, including those defined by three-layer residual networks with smooth activations, without any distributional assumption. At the same time, we prove there are simple functions in this class such that with the same number of training examples, the test error obtained by neural networks can be *much smaller* than *any* kernel method, including neural tangent kernels (NTK).

The main intuition is that *multi-layer* neural networks can implicitly perform hierarchical learning using different layers, which reduces the sample complexity comparing to "one-shot" learning algorithms such as kernel methods. In a follow-up work [2], this theory of hierarchical learning is further strengthened to incorporate the "backward feature correction" process when training deep networks.

In the end, we also prove a computation complexity advantage of ResNet with respect to other learning methods including linear regression over arbitrary feature mappings.

---

[*]V1 appears on this date, V2 slightly improved the lower bound, V3 strengthens experiments and adds citation to "backward feature correction" which is an even stronger form of hierarchical learning [2]. We would like to thank Greg Yang for many enlightening conversations as well as discussions on neural tangent kernels. A 45-min presentation of this result at the UC Berkeley Simons Institute can be found at https://youtu.be/NNPCk2gvTnI.

# 1 Introduction

Neural network learning has become a key practical machine learning approach and has achieved remarkable success in a wide range of real-world domains, such as computer vision, speech recognition, and game playing [19, 20, 23, 35]. On the other hand, from a theoretical standpoint, it is less understood that how large-scale, non-convex, non-smooth neural networks can be optimized efficiently over the training data and *generalize* to the test data with *relatively few* training examples.

There has been a sequence of research trying to address this question, showing that under certain conditions neural networks can be learned efficiently [3, 9–11, 16, 17, 22, 24, 25, 27–29, 36–39, 41, 44]. These provable guarantees typically come with strong assumptions and the proofs heavily rely on them. One common assumption from them is on the *input distribution*, usually being random Gaussian or sufficiently close to Gaussian. While providing great insights to the optimization side of neural networks, it is not clear whether these works emphasizing on Gaussian inputs can coincide with the neural network learning process in practice. Indeed, in nearly all real world data where deep learning is applied to, the input distributions are not close to Gaussians; even worse, there may be no simple model to capture such distributions.

The difficulty of modeling real-world distributions brings us back to the traditional PAC-learning language which is *distribution-free*. In this language, one of the most popular, *provable* learning methods is the kernel methods, defined with respect to kernel functions $K(x, x')$ over pairs of data $(x, x')$. The optimization task associated with kernel methods is convex, hence the convergence rate and the generalization error bound are well-established in theory.

Recently, there is a line of work studying the convergence of neural networks in the PAC-learning language, especially for over-parameterized neural networks [1, 4–8, 13–15, 21, 26, 45], putting neural network theory back to the distribution-free setting. Most of these works rely on the so-called Neural Tangent Kernel (NTK) technique [13, 21], by relating the training process of sufficiently over-parameterized (or even infinite-width) neural networks to the learning process over a kernel whose features are defined by the randomly initialized weights of the neural network. In other words, on the same training data set, these works prove that neural networks can efficiently learn a concept class with as good generalization as kernels, but *nothing more* is known.[1]

In contrast, in many practical tasks, neural networks give much better generalization error compared to kernels, although both methods can achieve zero training error. For example, ResNet achieves 96% test accuracy on the CIFAR-10 data set, but NTKs achieve 77% [7] and random feature kernels achieve 85% [33]. This gap becomes larger on more complicated data sets.

To separate the generalization power of neural networks from kernel methods, the recent work [40] tries to identify conditions where the solutions found by neural networks provably generalize better than kernels. This approach assumes that the optimization converges to minimal complexity solutions (i.e. the ones minimizing the value of the regularizer, usually the sum of squared Frobenius norms of weight matrices) of the training ob-

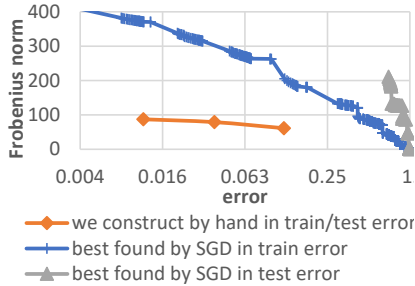

Figure 1: $d = 40, N = 5000$, after exhaustive search in network size, learning rate, weight decay, randomly initialized SGD still cannot find solutions with Frobenius norm comparable to what we construct by hand. Details and more experiments in Section 8.2.

---
[1]Technically speaking, the three-layer learning theorem of [4] is beyond NTK, because the learned weights across different layers interact with each other, while in NTK the learned weights of each layer only interact with random weights of other layers. However, there exist other kernels— such as recursive kernels [43] — that can more or less efficiently learn the same concept class proposed in [4].

jective. However, for most practical applications, it is unclear how, when training neural networks, minimal complexity solutions can be found efficiently by local search algorithms such as stochastic gradient descent. In fact, it is not true even for rather simple problems (see Figure 1).[2] Towards this end, the following fundamental question is largely unsolved:

*Can neural networks efficiently and distribution-freely learn a concept class, with better generalization than kernel methods?*

In this paper, we give arguably the *first* positive answer to this question for neural networks with ReLU activations. We show without any distributional assumption, a three-layer residual network (ResNet) can (improperly) learn a concept class that includes three-layer ResNets of smaller size and smooth activations. This learning process can be efficiently done by stochastic gradient descent (SGD), and the generalization error is also small if polynomially many training examples are given.

More importantly, we give a *provable separation* between the generalization error obtained by neural networks and *arbitrary kernel methods.* For some $\delta \in (0,1)$, with $N = O(\delta^{-2})$ training samples, we prove that neural networks can *efficiently* achieve generalization error $\delta$ for this concept class over *any distribution*; in contrast, there exists rather simple distributions such that any kernel method (including NTK, recursive kernel, etc) cannot have generalization error better than $\sqrt{\delta}$ for this class. To the best of our knowledge, this is the first work that gives provable, *efficiently achievable* separation between neural networks with ReLU activations and kernels in the distribution-free setting. In the end, we also prove a computation complexity advantage of neural networks with respect to linear regression over arbitrary feature mappings as well.

**Roadmap.** We present detailed overview of our positive and negative results in Section 2 and 3. Then, we introduce notations in Section 4, formally define our concept class in Section 5, and give proof overviews in Section 6 and 7.

## 2 Positive Result: The Learnability of Three-Layer ResNet

In this paper, we consider *learner networks* that are single-skip three-layer ResNet with ReLU activation, defined as a function $\mathsf{out} \colon \mathbb{R}^d \to \mathbb{R}^k$:

$$\mathsf{out}(x) = \mathbf{A}\left(\sigma\left(\mathbf{W}x + b_1\right) + \sigma\left(\mathbf{U}\sigma\left(\mathbf{W}x + b_1\right) + b_2\right)\right) \tag{2.1}$$

Here, $\sigma$ is the ReLU function, $\mathbf{W} \in \mathbb{R}^{m \times d}$ and $\mathbf{U} \in \mathbb{R}^{m \times m}$ are the hidden weights, $\mathbf{A} \in \mathbb{R}^{k \times m}$ is the output weight, and $b_1, b_2 \in \mathbb{R}^m$ are two bias vectors.

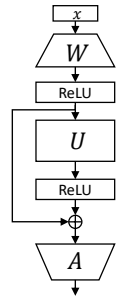

We wish to learn a concept class given by target functions that can be written as

$$\mathcal{H}(x) = \mathcal{F}(x) + \alpha \mathcal{G}\left(\mathcal{F}(x)\right) \tag{2.2}$$

where $\alpha \in [0,1)$ and $\mathcal{G} \colon \mathbb{R}^k \to \mathbb{R}^k, \mathcal{F} \colon \mathbb{R}^d \to \mathbb{R}^k$ are two functions that can be written as two-layer networks with smooth activations (see Section 5 for the formal definition). Intuitively, the target function is a mixture of two parts: the base signal $\mathcal{F}$, which is simpler and contributes more to the target, and the composite signal $\mathcal{G}(\mathcal{F})$, which is more complicated but contributes less. As an analogy, $\mathcal{F}$ could capture the signal in which "85%" examples in CIFAR-10 can be learned by kernel methods, and $\mathcal{G}(\mathcal{F})$ could capture the additional "11%" examples that are more complicated. The goal is to use three-layer ResNet (2.1) to *improperly* learn this concept class (2.2), meaning learning "both" the base and composite signals, with as few samples as possible. In this paper, we consider a simple $\ell_2$ regression task where the features $x \in \mathbb{R}^d$ and labels $y \in \mathbb{R}^k$ are sampled from some unknown distribution $\mathcal{D}$. Thus, given a network $\mathsf{out}(x)$, the population risk is

$$\mathbb{E}_{(x,y)\sim\mathcal{D}} \frac{1}{2} \|\mathsf{out}(x) - y\|_2^2 .$$

---
[2]Consider the class of degree-6 polynomials over 6 coordinates of the $d$-dimensional input. There exist two-layer networks with F-norm $O(\sqrt{d})$ implementing this function (thus have near-zero training and testing error). By Rademacher complexity, $O(d)$ samples suffice to learn if we are able to find a *minimal complexity* solution. Unfortunately, due to the non-convexity of the optimization landscape, two-layer networks can not be trained to match this F-norm even with $O(d^2)$ samples, see Figure 1.

To illustrate our result, we first assume *for simplicity* that $y = \mathcal{H}(x)$ for some $\mathcal{H}$ of the form (2.2) (so the optimal target has zero regression error). Our main theorem can be sketched as follows.

Let $C_\mathcal{F}$ and $C_\mathcal{G}$ respectively be the individual "complexity" of $\mathcal{F}$ and $\mathcal{G}$, which at a high level, capture the size and smoothness of $\mathcal{F}$ and $\mathcal{G}$. This complexity notion shall be formally introduced in Section 4, and is used by prior works such as [4, 8, 43].

**Theorem 1** (ResNet, sketched). *For any distribution over $x$, for every $\delta \in \left((\alpha C_\mathcal{G})^4, 1\right)$, with probability at least $0.99$, SGD efficiently learns a network $\mathsf{out}(x)$ in the form (2.1) satisfying*

$$\mathbb{E}_{(x,y)\sim\mathcal{D}} \frac{1}{2} \|\mathsf{out}(x) - y\|_2^2 \leq \delta \quad using \quad N = \widetilde{O}\left(\frac{C_\mathcal{F}^2}{\delta^2}\right) \; samples$$

*The running time of SGD is polynomial in* $\mathsf{poly}(C_\mathcal{G}, C_\mathcal{F}, \alpha^{-1})$.

In other words, ResNet is capable of achieving population risk $\alpha^4$, or equivalently learning the output $\mathcal{H}(x)$ up to $\alpha^2$ error. In our full theorem, we also allow label $y$ to be generated from $\mathcal{H}(x)$ with error, thus our result also holds in the agnostic learning framework.

## 2.1 Our Contributions

Our main contribution is to obtain time and sample complexity in $C_\mathcal{F}$ and $C_\mathcal{G}$ *without any dependency* on the composed function $\mathcal{G}(\mathcal{F})$ as in prior work [4, 43]. We illustrate this crucial difference with an example. Suppose $x \sim \mathcal{N}(0, \mathbf{I}/d)$, $k = 2$ and $\mathcal{F} \in \mathbb{R}^d \to \mathbb{R}^2$ consists of two linear function: $\mathcal{F}(x) = \left(\langle w_1^*, x\rangle, \langle w_2^*, x\rangle\right)$ with $\|w_1^*\|_2, \|w_2^*\|_2 = \sqrt{d}$, and $\mathcal{G}$ is degree-10 polynomial with constant coefficient. As we shall see, $C_\mathcal{F} = O(\sqrt{d})$ and $C_\mathcal{G} = \widetilde{O}(1)$. Theorem 1 implies

- we need $\widetilde{O}(d)$ samples to efficiently learn $\mathcal{H} = \mathcal{F} + \alpha \mathcal{G}(\mathcal{F})$ up to accuracy $\widetilde{O}(\alpha^2)$.

In contrast, the complexity of $\mathcal{G}(\mathcal{F})$ is $\widetilde{O}((\sqrt{d})^{10})$, so

- prior works [4, 43] need $\widetilde{\Omega}(d^{10})$ samples to *efficiently* learn $\mathcal{H}$ up to any accuracy $o(\alpha)$,

even if $\mathcal{G}(x)$ is of some simple form such as $\langle w_1^*, x\rangle^{10} - \langle w_2^*, x\rangle^{10}$.[3]

**Inductive Bias.** Our network is over-parameterized, thus intuitively in the example above, with only $O(d)$ training examples, the learner network could over-fit to the training data since it has to decide from a set of $d^{10}$ many possible coefficients to learn the degree 10 polynomial $\mathcal{G}$. This is indeed the case if we learn the target function using kernels, or possibly even learn it with a two-layer network. However, three-layer ResNet posts a *completely different* inductive bias, and manages to avoid over-fitting to $\mathcal{G}(\mathcal{F})$ with the help from $\mathcal{F}$.

---

[3]Of course, if one knew a priori the form $\mathcal{H}(x) = \langle w_1^*, x\rangle^{10} - \langle w_2^*, x\rangle^{10}$, one could also try to solve it directly by minimizing objective $(\langle w_1^*, x\rangle^{10} - \langle w_2^*, x\rangle^{10} + \langle w_2, x\rangle^{10} - \langle w_1, x\rangle^{10})^2$ over $w_1, w_2 \in \mathbb{R}^d$. Unfortunately, the underlying optimization process is highly non-convex and it remains unclear how to minimize it efficiently. Using matrix sensing [29], one can efficiently learn such $\mathcal{H}(x)$ in sample complexity $\widetilde{O}(d^5)$.

**Implicit Hierarchical Learning using Forward Feature Learning.** Since $\mathcal{H}(x) = \mathcal{F}(x) + \alpha \mathcal{G}(\mathcal{F}(x))$, if we only learn $\mathcal{F}$ but not $\mathcal{G}(\mathcal{F})$, we will have regression error $\approx (\alpha C_\mathcal{G})^2$. Thus, to get to regression error $(\alpha C_\mathcal{G})^4$, Theorem 1 shows that ResNet is also capable of learning $\mathcal{G}(\mathcal{F})$ up to some good accuracy with *relatively few* training examples. This is also observed in practice, where with this number of training examples, three-layer fully-connected networks and kernel methods can indeed fail to learn $\mathcal{G}(\mathcal{F})$ up to any non-trivial accuracy, see Figure 2.

Intuitively, there is a hierarchy of the learning process: we would like to first learn $\mathcal{F}$, and then we could learn $\mathcal{G}(\mathcal{F})$ much easier with the help of $\mathcal{F}$ using the *residual link*. In our learner network (2.1), the first hidden layer serves to learn $\mathcal{F}$ and the second hidden layer serves to learn $\mathcal{G}$ with the help of $\mathcal{F}$, which reduces the sample complexity. However, the important message is that $\mathcal{F}$ and $\mathcal{G}$ are *not* given as separate data to the network, rather the learning algorithm has to *disentangle* them from the "combined" function $\mathcal{H} = \mathcal{F} + \alpha \mathcal{G}(\mathcal{F})$ automatically during the training process. Moreover, since we *train both layers simultaneously*, the learning algorithm also has to *distribute* the learning task of $\mathcal{F}$ and $\mathcal{G}$ onto different layers automatically. We call this process "forward feature learning":

> **Hierarchical Learning in ResNet: The Forward Feature Learning**
>
> During the training process of a residual network, the lower-level layers *automatically* learn an approximation of the lower-complexity features/signals in the target function. It then **forward** these features to the higher-level layers in the network to *further* learn the higher-complexity features/signals in the target function.

We point out forward feature learning is different from layer-wise training. For instance, our result cannot be obtained by first training the hidden layer close to the input, and then fixing it and training the hidden layer close to the output. Since it could be the case the first layer incurs some $\alpha$ error (since it cannot learn $\mathcal{G}(\mathcal{F})$ directly), then it could be really hard, or perhaps impossible, for the second layer to fix it only using inputs of the form $\mathcal{F}(x) \pm \alpha$. In other words, it is crucial that the two hidden layers are *simultaneously trained*. [4]

**A follow-up work.** In a follow-up work [2], this theory of hierarchical learning is *strengthened* to further incorporate the **backward feature correction** step when training deep neural networks. In the language of this paper, when the two layers trained together, given enough samples, the accuracy in the first layer can actually be *improved* from $\mathcal{F} \pm \alpha$ to *arbitrarily close to* $\mathcal{F}$ during the training process. As a consequence, the final training and generalization error can be arbitrarily small as well, as opposite to $\alpha^2$ (or equivalently population risk $\alpha^4$) in this work. The new "backward feature correction" is also critical to extend the hierarchical learning process from 3 layers to *arbitrarily number of layers*.

## 3 Negative Results

### 3.1 Limitation of Kernel Methods

Given (Mercer) kernels $K_1, \ldots, K_k : \mathbb{R}^{d \times d} \to \mathbb{R}$ and training examples $\{(x^{(i)}, y^{(i)})\}_{i \in [N]}$ from $\mathcal{D}$, a kernel method tries to learn a function $\mathfrak{K} : \mathbb{R}^d \to \mathbb{R}^k$ where each

$$\mathfrak{K}_j(x) = \sum_{n \in [N]} K_j(x, x^{(n)}) \cdot w_{j,n} \tag{3.1}$$

---

[4] This does not mean that the error of the first layer can be reduced by its own, since it is still possible for the first layer to learn $\mathcal{F} + \alpha \mathcal{R}(x) \pm \alpha^2$ and the second layer to learn $\mathcal{G}(\mathcal{F})(x) - \mathcal{R}(x) \pm \alpha$, for an arbitrary (bounded) function $\mathcal{R}$.

is parameterized by a weight vector $w_j \in \mathbb{R}^N$. Usually, for the $\ell_2$ regression task, a kernel method finds the optimal weights $w_1, \ldots, w_k \in \mathbb{R}^N$ by solving the following convex minimization problem

$$\frac{1}{N} \sum_{i=1}^{N} \sum_{j \in [k]} \left( \sum_{n \in [N]} K_j(x^{(i)}, x^{(n)}) w_{j,n} - y_j^{(i)} \right)^2 + R(w) \quad (3.2)$$

for some convex regularizer $R(w)$.[5] In this paper, however, we do not make assumptions about how $\mathfrak{K}(x)$ is found as the optimal solution of the training objective. Instead, we focus on *any* kernel regression function that can be written in the form (3.1).

Most of the widely-used kernels are Mercer kernels.[6] This includes (1) *Gaussian kernel* $K(x,y) = e^{-\|x-y\|_2^2/h}$; (2) *arcsin kernel* $K(x,y) = \arcsin\left(\langle x, y \rangle / (\|x\|_2 \|y\|_2)\right)$; (3) *recursive kernel* with any recursive function [43]; (4) *random feature kernel* $K(x,y) = \mathbb{E}_{w \sim \mathcal{W}} \phi_w(x) \phi_w(y)$ for any function $\phi_w(\cdot)$ and distribution $\mathcal{W}$; (5) the *conjugate kernel* defined by the last hidden layer of random initialized neural networks [12]; (6) the *neural tangent kernels (NTK)* for fully-connected [21] networks, convolutional networks [7, 42] or more generally for any architectures [42].

Our theorem can be sketched as follows:

**Theorem 2** (kernel, sketched). *For every constant $k \geq 2$, for every sufficiently large $d \geq 2$, there exist concept classes consisting of functions $\mathcal{H}(x) = \mathcal{F}(x) + \alpha \mathcal{G}(\mathcal{F}(x))$ with complexities $C_{\mathcal{F}}, C_{\mathcal{G}}$ and $\alpha \in (0, \frac{1}{C_{\mathcal{G}}})$ such that, letting*

$N_{\mathsf{res}}$ *be the sample complexity from Theorem 1 to achieve $\alpha^{3.9}$ population risk,*

*then there exists simple distributions $\mathcal{D}$ over $(x, \mathcal{H}(x))$ such that, for at least 99% of the functions $\mathcal{H}$ in this concept class, even given $N = O\bigl((N_{\mathsf{res}})^{k/2}\bigr)$ training samples from $\mathcal{D}$, any function $\mathfrak{K}(x)$ of the form (3.1) has to suffer population risk*

$$\mathbb{E}_{(x,y) \sim \mathcal{D}} \tfrac{1}{2} \|\mathfrak{K}(x) - y\|_2^2 > \alpha^2 \quad \text{even if the label } y = \mathcal{H}(x) \text{ has no error.}$$

**Contribution and Intuition.** Let us compare this to Theorem 1. While both algorithms are efficient, neural networks (trained by SGD) achieve population risk $\alpha^{3.9}$ using $N_{\mathsf{res}}$ samples for *any* distribution over $x$, while kernel methods cannot achieve any population risk better than $\alpha^2$ for some simple distributions even with $N = (N_{\mathsf{res}})^{k/2} \gg N_{\mathsf{res}}$ samples.[7] Our two theorems together gives a provable separation between the generalization error of the solutions found by neural networks and kernel methods, in the *efficiently computable regime*.

More specifically, recall $C_{\mathcal{F}}$ and $C_{\mathcal{G}}$ only depend on individual complexity of $\mathcal{G}, \mathcal{F}$, but not on $\mathcal{G}(\mathcal{F})$. In Theorem 2, we will construct $\mathcal{F}$ as linear functions and $\mathcal{G}$ as degree-$k$ polynomials. This ensures $C_{\mathcal{F}} = O(\sqrt{d})$ and $C_{\mathcal{G}} = O(1)$ for $k$ being constant, but the combined complexity of $\mathcal{G}(\mathcal{F})$ is as high as $\Omega(d^{k/2})$. Since ResNet can perform hierarchical learning, it only needs sample complexity $N_{\mathsf{res}} = O(d/\alpha^8)$ instead of paying (square of) the combined complexity $\Omega(d^k)$.

In contrast, a kernel method is not hierarchical: rather than discovering $\mathcal{F}$ first and then learning $\mathcal{G}(\mathcal{F})$ with the guidance of it, kernel method tries to learn everything *in one shot*. This unavoidably requires the sample complexity to be at least $\Omega(d^k)$. Intuitively, as the kernel method tries to learn $\mathcal{G}(\mathcal{F})$ from scratch, this means that it has to take into account all $\Omega(d^k)$ many possible choices of $\mathcal{G}(\mathcal{F})$ (recall that $\mathcal{G}$ is a degree $k$ polynomial over dimension $d$). On the other hand, a kernel method with $N$ samples only has $N$-degrees of freedom (for each output dimension). This means, if $N \ll o(d^k)$, kernel method simply does not have enough degrees of freedom to distinguish between different $\mathcal{G}(\mathcal{F})$, so has to pay $\Omega(\alpha^2)$ in population risk. Choosing for instance $\alpha = d^{-0.1}$, we have the desired negative result for all $N \leq O\big((N_{\mathsf{res}})^{k/2}\big) \ll o(d^k)$.

---

[5]In many cases, $R(w) = \lambda \cdot \sum_{j \in [k]} w_j^\top K_j w_j$ is the norm associated with the kernel, for matrix $K_j \in \mathbb{R}^{N \times N}$ defined as $[K_j]_{i,n} = K_j(x^{(i)}, x^{(n)})$.

[6]Recall a Mercer kernel $K \colon \mathbb{R}^{d \times d} \to \mathbb{R}$ can be written as $K(x,y) = \langle \phi(x), \phi(y) \rangle$ where $\phi \colon \mathbb{R}^d \to \mathcal{V}$ is a feature mapping to some inner product space $\mathcal{V}$.

[7]It is necessary the negative result of kernel methods is distribution dependent, since for trivial distributions where $x$ is non-zero only on the first constantly many coordinates, both neural networks and kernel methods can learn it with constantly many samples.

## 3.2 Limitation of Linear Regression Over Feature Mappings

Given an arbitrary feature mapping $\phi \colon \mathbb{R}^d \to \mathbb{R}^D$, one may consider learning a linear function over $\phi$. Namely, to learn a function $\mathfrak{F} \colon \mathbb{R}^d \to \mathbb{R}^k$ where each

$$\mathfrak{F}_j(x) = w_j^\top \phi(x) \tag{3.3}$$

is parameterized by a weight vector $w_j \in \mathbb{R}^D$. Usually, these weights are determined by minimizing the following regression objective:[8]

$$\tfrac{1}{N} \sum_{i \in [N]} \sum_{j \in [k]} \left( w_j^\top \phi(x^{(i)}) - y_j^{(i)} \right)^2 + R(w)$$

for some regularizer $R(w)$. In this paper, we do not make assumptions about how the weighted are found. Instead, we focus on *any* linear function over such feature mapping in the form (3.3).

**Theorem 3** (feature mapping, sketched)**.** *For sufficiently large integers $d, k$, there exist concept classes consisting of functions $\mathcal{H}(x) = \mathcal{F}(x) + \alpha \mathcal{G}\left(\mathcal{F}(x)\right)$ with complexities $C_\mathcal{F}, C_\mathcal{G}$ and $\alpha \in (0, \frac{1}{C_\mathcal{G}})$ such that, letting*

*$T_{\mathsf{res}}$ be the time complexity from Theorem 1 to achieve $\alpha^{3.9}$ population risk,*

*then for at least 99% of the functions $\mathcal{H}$ in this concept class, even with arbitrary $D = (T_{\mathsf{res}})^2$ dimensional feature mapping, any function $\mathfrak{F}(x)$ of the form (3.3) has to suffer population risk*

$$\mathbb{E}_{(x,y) \sim \mathcal{D}} \tfrac{1}{2} \left\| \mathfrak{F}(x) - y \right\|_2^2 > \alpha^2 \quad \text{even if the label } y = \mathcal{H}(x) \text{ has no error.}$$

**Interpretation.** Since any algorithm that optimizes linear functions over $D$-dimensional feature mapping has to run in time $\Omega(D)$, this proves a time complexity separation between neural networks (say, for achieving population risk $\alpha^{3.9}$) and linear regression over feature mappings (for achieving even any population risk better than $\alpha^2 \gg \alpha^{3.9}$). Usually, such an algorithm also has to suffer from $\Omega(D)$ space complexity. If that happens, we also have a space complexity separation. Our hard instance in proving Theorem 3 is the same as Theorem 2, and the proof is analogous.

## 4 Notations

We denote by $\|w\|_2$ and $\|w\|_\infty$ the Euclidean and infinity norms of vectors $w$, and $\|w\|_0$ the number of non-zeros of $w$. We also abbreviate $\|w\| = \|w\|_2$ when it is clear from the context. We denote the row $\ell_p$ norm for $\mathbf{W} \in \mathbb{R}^{m \times d}$ (for $p \geq 1$) as

$$\|\mathbf{W}\|_{2,p} \stackrel{\text{def}}{=} \big( \sum_{i \in [m]} \|w_i\|_2^p \big)^{1/p}.$$

By definition, $\|\mathbf{W}\|_{2,2} = \|\mathbf{W}\|_F$ is the Frobenius norm of $\mathbf{W}$. We use $\|\mathbf{W}\|_2$ to denote the matrix spectral norm. For a diagonal matrix $D$ we use $\|D\|_0$ to denote its sparsity. For a matrix $\mathbf{W} \in \mathbb{R}^{m \times d}$, we use $\mathbf{W}_i$ or $w_i$ to denote the $i$-th row of $\mathbf{W}$.

We use $\mathcal{N}(\mu, \sigma)$ to denote Gaussian distribution with mean $\mu$ and variance $\sigma$; or $\mathcal{N}(\mu, \Sigma)$ to denote Gaussian vector with mean $\mu$ and covariance $\Sigma$. We use $\mathbb{1}_{event}$ or $\mathbb{1}[event]$ to denote the indicator function of whether *event* is true. We use $\sigma(\cdot)$ to denote the ReLU function, namely $\sigma(x) = \max\{x, 0\} = \mathbb{1}_{x \geq 0} \cdot x$. Given univariate function $f \colon \mathbb{R} \to \mathbb{R}$, we also use $f$ to denote the same function over vectors: $f(x) = (f(x_1), \ldots, f(x_m))$ if $x \in \mathbb{R}^m$.

---
[8]If $R(w)$ is the $\ell_2$ regularizer, then this becomes a kernel method again since the minimizer can be written in the form (3.1). For other regularizers, this may not be the case.

For notation simplicity, throughout this paper "with high probability" (or w.h.p.) means with probability $1 - e^{-c \log^2 m}$ for a sufficiently large constant $c$. We use $\widetilde{O}$ to hide $\mathsf{polylog}(m)$ factors.

**Function complexity.** The following notions introduced in [4] measure the complexity of any infinite-order smooth function $\phi \colon \mathbb{R} \to \mathbb{R}$. Suppose $\phi(z) = \sum_{i=0}^{\infty} c_i z^i$ is its Taylor expansion. [9]

$$\mathfrak{C}_\varepsilon(\phi) = \mathfrak{C}_\varepsilon(\phi, 1) \stackrel{\mathrm{def}}{=} \sum_{i=0}^{\infty} \left( (C^*)^i + \left( \frac{\sqrt{\log(1/\varepsilon)}}{\sqrt{i}} C^* \right)^i \right) |c_i|$$

$$\mathfrak{C}_\mathfrak{s}(\phi) = \mathfrak{C}_\mathfrak{s}(\phi, 1) \stackrel{\mathrm{def}}{=} C^* \sum_{i=0}^{\infty} (i+1) |c_i|$$

where $C^*$ is a sufficiently large constant (e.g., $10^4$).

*Example* 4.1. If $\phi(z) = e^{c \cdot z} - 1$, $\sin(c \cdot z)$, $\cos(c \cdot z)$ or degree-$c$ polynomial for constant $c$, then $\mathfrak{C}_\varepsilon(\phi, 1) = o(1/\varepsilon)$ and $\mathfrak{C}_\mathfrak{s}(\phi, 1) = O(1)$. If $\phi(z) = \mathsf{sigmoid}(z)$ or $\tanh(z)$, to get $\varepsilon$ approximation we can truncate their Taylor series at degree $\Theta(\log \frac{1}{\varepsilon})$. One can verify that $\mathfrak{C}_\varepsilon(\phi, 1) \leq \mathsf{poly}(1/\varepsilon)$ by the fact that $(\log(1/\varepsilon)/i)^i \leq \mathsf{poly}(\varepsilon^{-1})$ for every $i \geq 1$, and $\mathfrak{C}_\mathfrak{s}(\phi, 1) \leq O(1)$.

## 5 Concept Class

We consider learning some unknown distribution $\mathcal{D}$ of data points $z = (x, y) \in \mathbb{R}^d \times \mathbb{R}^k$, where $x \in \mathbb{R}^d$ is the input vector and $y$ is the associated label. Let us consider *target functions* $\mathcal{H} \colon \mathbb{R}^d \to \mathbb{R}^k$ coming from the following concept class.

**Concept 1.** $\mathcal{H}$ *is given by two smooth functions* $\mathcal{F}, \mathcal{G} \colon \mathbb{R}^k \to \mathbb{R}^k$ *and a value* $\alpha \in \mathbb{R}_+$:

$$\mathcal{H}(x) = \mathcal{F}(x) + \alpha \mathcal{G}(\mathcal{F}(x)) \quad, \tag{5.1}$$

*where for each output coordinate* $r$,

$$\mathcal{F}_r(x) = \sum_{i \in [p_\mathcal{F}]} a^*_{\mathcal{F},r,i} \cdot \mathcal{F}_{r,i}(\langle w^*_{r,i}, x \rangle) \quad \text{and} \quad \mathcal{G}_r(h) = \sum_{i \in [p_\mathcal{G}]} a^*_{\mathcal{G},r,i} \cdot \mathcal{G}_{r,i}(\langle v^*_{r,i}, h \rangle) \tag{5.2}$$

*for some parameters* $a^*_{\mathcal{F},r,i}, a^*_{\mathcal{G},r,i} \in [-1, 1]$ *and vectors* $w^*_{r,i} \in \mathbb{R}^d$ *and* $v^*_{r,i} \in \mathbb{R}^k$. *We assume for simplicity* $\|w^*_{r,i}\|_2 = \|v^*_{r,i}\|_2 = 1/\sqrt{2}$.[10] *For simplicity, we assume* $\|x\|_2 = 1$ *and* $\|\mathcal{F}(x)\|_2 = 1$ *for* $(x, y) \sim \mathcal{D}$ *and in Appendix A we state a more general Concept 2 without these assumptions.*[11]

We denote by $\mathfrak{C}_\varepsilon(\mathcal{F}) = \max_{r,i} \{\mathfrak{C}_\varepsilon(\mathcal{F}_{r,i})\}$ and $\mathfrak{C}_\mathfrak{s}(\mathcal{F}) = \max_{r,i} \{\mathfrak{C}_\mathfrak{s}(\mathcal{F}_{r,i})\}$. Intuitively, $\mathcal{F}$ and $\mathcal{G}$ are both generated by two-layer neural networks with smooth activation functions $\mathcal{F}_{r,i}$ and $\mathcal{G}_{r,i}$.

Borrowing the *agnostic PAC-learning language*, our concept class consists of all functions $\mathcal{H}(x)$ in the form of Concept 1 with complexity bounded by tuple $(p_F, C_F, p_G, C_G)$. Let $\mathsf{OPT}$ be the population risk achieved by the *best* target function in this concept class. Then, our goal is to learn this concept class with population risk $O(\mathsf{OPT}) + \varepsilon$ using sample and time complexity *polynomial* in $p_F, C_F, p_G, C_G$ and $1/\varepsilon$. In the remainder of this paper, to simplify notations, we do not explicitly define this concept class parameterized by $(p_F, C_F, p_G, C_G)$. Instead, we equivalently state our theorem with respect to any (unknown) fixed target function $\mathcal{H}$ with with population risk $\mathsf{OPT}$:

$$\mathbb{E}_{(x,y)\sim\mathcal{D}}\left[\tfrac{1}{2}\|\mathcal{H}(x) - y\|_2^2\right] \leq \mathsf{OPT} \ .$$

---
[9]In [4, ver.5], they have used $(i+1)^{1.75}|c_i|$ instead of $(i+1)|c_i|$. For the purpose of this paper we have tightened this complexity measure.

[10]For general $\|w^*_{1,i}\|_2 \leq B, \|w^*_{2,i}\|_2 \leq B, |a^*_{r,i}| \leq B$, the scaling factor $B$ can be absorbed into the activation function $\phi'(x) = \phi(Bx)$. Our results then hold by replacing the complexity of $\phi$ with $\phi'$.

[11]Since we use ReLU networks as learners, they are positive homogeneous so to learn general functions $\mathcal{F}, \mathcal{G}$ which may not be positive homogenous, it is in some sense necessary that the inputs are scaled properly.

In the analysis we adopt the following notations. For every $(x,y) \sim \mathcal{D}$, it satisfies $\|\mathcal{F}(x)\|_2 \leq \mathfrak{B}_\mathcal{F}$ and $\|\mathcal{G}(\mathcal{F}(x))\|_2 \leq \mathfrak{B}_{\mathcal{F}\circ\mathcal{G}}$. We assume $\mathcal{G}(\cdot)$ is $\mathfrak{L}_\mathcal{G}$-Lipschitz continuous. It is a simple exercise (see Fact A.3) to verify that $\mathfrak{L}_\mathcal{G} \leq \sqrt{k}p_\mathcal{G}\mathfrak{C}_\mathfrak{s}(\mathcal{G})$, $\mathfrak{B}_\mathcal{F} \leq \sqrt{k}p_\mathcal{F}\mathfrak{C}_\mathfrak{s}(\mathcal{F})$ and $\mathfrak{B}_{\mathcal{F}\circ\mathcal{G}} \leq \mathfrak{L}_\mathcal{G}\mathfrak{B}_\mathcal{F} + \sqrt{k}p_\mathcal{G}\mathfrak{C}(\mathcal{G}) \leq kp_\mathcal{F}\mathfrak{C}_\mathfrak{s}(\mathcal{F})p_\mathcal{G}\mathfrak{C}_\mathfrak{s}(\mathcal{G})$.

## 6 Overview of Theorem 1

We learn the unknown distribution $\mathcal{D}$ with three-layer ResNet with ReLU activation (2.1) as learners. For notation simplicity, we absorb the bias vector into weight matrix: that is, given $\mathbf{W} \in \mathbb{R}^{m\times d}$ and bias $b_1 \in \mathbb{R}^m$, we rewrite $\mathbf{W}x + b$ as $\mathbf{W}(x,1)$ for a new weight matrix $\mathbf{W} \in \mathbb{R}^{m\times(d+1)}$. We also re-parameterize $\mathbf{U}$ as $\mathbf{U} = \mathbf{V}\mathbf{A}$ and we find this parameterization (similar to the "bottleneck" structure in ResNet) simplifies the proof and also works well empirically for our concept class. After such notation simplification and re-parameterization, we can rewrite $\mathsf{out}(x)\colon \mathbb{R}^d \to \mathbb{R}^k$ as

$$\mathsf{out}(\mathbf{W},\mathbf{V};x) = \mathsf{out}(x) = \mathsf{out}_1(x) + \mathbf{A}\sigma\left((\mathbf{V}^{(0)} + \mathbf{V})(\mathsf{out}_1(x),1)\right)$$

$$\mathsf{out}_1(\mathbf{W},\mathbf{V};x) = \mathsf{out}_1(x) = \mathbf{A}\sigma(\mathbf{W}^{(0)} + \mathbf{W})(x,1) \ .$$

Above, $\mathbf{A} \in \mathbb{R}^{k\times m}, \mathbf{V}^{(0)} \in \mathbb{R}^{m\times(k+1)}, \mathbf{W}^{(0)} \in \mathbb{R}^{m\times(d+1)}$ are weight matrices corresponding to random initialization, and $\mathbf{W} \in \mathbb{R}^{m\times(k+1)}, \mathbf{W} \in \mathbb{R}^{m\times(d+1)}$ are the additional weights to be learned by the algorithm. To prove the strongest result, we only train $\mathbf{W}, \mathbf{V}$ and do not train $\mathbf{A}$.[12] We consider random Gaussian initialization where the entries of $\mathbf{A}, \mathbf{W}^{(0)}, \mathbf{V}^{(0)}$ are independently generated as follows:

$$\mathbf{A}_{i,j} \sim \mathcal{N}\left(0, \tfrac{1}{m}\right) \qquad [\mathbf{W}^{(0)}]_{i,j} \sim \mathcal{N}\left(0, \sigma_w^2\right) \qquad [\mathbf{V}^{(0)}]_{i,j} \sim \mathcal{N}\left(0, \sigma_v^2/m\right)$$

In this paper we focus on the $\ell_2$ loss function between $\mathcal{H}$ and $\mathsf{out}$, given as:

$$\mathsf{Obj}(\mathbf{W},\mathbf{V};(x,y)) = \frac{1}{2}\|y - \mathsf{out}(\mathbf{W},\mathbf{V};x)\|_2^2 \tag{6.1}$$

We consider the vanilla SGD algorithm given in Algorithm 1.[13]

---

**Algorithm 1** SGD

1: Initially $\mathbf{W}_0, \mathbf{V}_0 = 0$.
2: **for** $t = 0, 1, \cdots, T-1$ **do**
3:     Sample $(x_t, y_t) \sim \mathcal{D}$.
4:     Define $\ell_2$ objective $\mathsf{Obj}(\mathbf{W},\mathbf{V};(x_t,y_t)) = \frac{1}{2}\|y_t - \mathsf{out}(\mathbf{W},\mathbf{V};x_t)\|_2^2$.
5:     Update $\mathbf{W}_{t+1} \leftarrow \mathbf{W}_t - \eta_w \frac{\partial \mathsf{Obj}(\mathbf{W},\mathbf{V};(x_t,y_t))}{\partial \mathbf{W}}\big|_{\mathbf{W}=\mathbf{W}_t, \mathbf{V}=\mathbf{V}_t}$.
6:     Update $\mathbf{V}_{t+1} \leftarrow \mathbf{V}_t - \eta_v \frac{\partial \mathsf{Obj}(\mathbf{W},\mathbf{V};(x_t,y_t))}{\partial \mathbf{V}}\big|_{\mathbf{W}=\mathbf{W}_t, \mathbf{V}=\mathbf{V}_t}$.
7: **end for**

---

[12]This can be more meaningful than training all the layers together, in which if one is not careful with parameter choices, the training process can degenerate as if only the last layer is trained [12]. (That is a convex kernel method.) Of course, as a simple corollary, our result also applies to training all the layers together, with appropriately chosen random initialization and learning rate.

[13]Performing SGD with respect to $\mathbf{W}^{(0)} + \mathbf{W}$ and $\mathbf{V}^{(0)} + \mathbf{V}$ is the *same* as that with respect to $\mathbf{W}$ and $\mathbf{V}$; we introduce $\mathbf{W}^{(0)}, \mathbf{V}^{(0)}$ notation for analysis purpose. Note also, one can alternatively consider having a training set and then performing SGD on this training set with multiple passes; similar results can be obtained.

**Theorem 1.** *Under Concept 1 or Concept 2, for every $\alpha \in \left(0, \widetilde{\Theta}(\frac{1}{kp_{\mathcal{G}}\mathfrak{C}_{\mathfrak{s}}(\mathcal{G})})\right)$ and $\delta \geq \mathsf{OPT} + \widetilde{\Theta}\left(\alpha^4 (kp_{\mathcal{G}}\mathfrak{C}_{\mathfrak{s}}(\mathcal{G}))^4 (1 + \mathfrak{B}_{\mathcal{F}})^2\right)$. There exist $M = \mathsf{poly}(\mathfrak{C}_{\alpha}(\mathcal{F}), \mathfrak{C}_{\alpha}(\mathcal{G}), p_{\mathcal{F}}, \alpha^{-1})$ satisfying that for every $m \geq M$, with high probability over $\mathbf{A}, \mathbf{W}^{(0)}, \mathbf{V}^{(0)}$, for a wide range of random initialization parameters $\sigma_w, \sigma_v$ (see Table 1), choosing*

$$T = \widetilde{\Theta}\left(\frac{(kp_{\mathcal{F}}\mathfrak{C}_{\mathfrak{s}}(\mathcal{F}))^2}{\min\{1, \delta^2\}}\right) \quad \eta_w = \widetilde{\Theta}\left(\min\{1, \delta\}\right) \quad \eta_v = \eta_w \cdot \widetilde{\Theta}\left(\frac{\alpha p_{\mathcal{G}}\mathfrak{C}_{\mathfrak{s}}(\mathcal{G})}{p_{\mathcal{F}}\mathfrak{C}_{\mathfrak{s}}(\mathcal{F})}\right)^2$$

*With high probability, the SGD algorithm satisfies*

$$\frac{1}{T}\sum_{t=0}^{T-1} \mathop{\mathbb{E}}_{(x,y)\sim\mathcal{D}} \|\mathcal{H}(x) - \mathsf{out}(\mathbf{W}_t, \mathbf{V}_t; x)\|_2^2 \leq O(\delta) \ .$$

As a corollary, under Concept 1, we can archive population risk

$$\frac{1}{T}\sum_{t=0}^{T-1} \mathop{\mathbb{E}}_{(x,y)\sim\mathcal{D}} \|\mathcal{H}(x) - \mathsf{out}(\mathbf{W}_t, \mathbf{V}_t; x)\|_2^2 \leq O(\mathsf{OPT}) + \widetilde{\Theta}\left(\alpha^4 (kp_{\mathcal{G}}\mathfrak{C}_{\mathfrak{s}}(\mathcal{G}))^4\right) \quad \text{using sample complexity } T \ .$$

(6.2)

*Remark* 6.1. Our Theorem 1 is almost in the PAC-learning language, except that the final error has an additive $\alpha^4$ term that can not be arbitrarily small.

## 6.1 Proof Overview

In the analysis, let us define diagonal matrices

$$D_{\mathbf{W}^{(0)}} = \mathbf{diag}\{\mathbb{1}_{\mathbf{W}^{(0)}(x,1)\geq 0}\} \qquad D_{\mathbf{V}^{(0)},\mathbf{W}} = \mathbf{diag}\{\mathbb{1}_{\mathbf{V}^{(0)}(\mathsf{out}_1(x),1)\geq 0}\}$$
$$D_{\mathbf{W}} = \mathbf{diag}\{\mathbb{1}_{(\mathbf{W}^{(0)}+\mathbf{W})(x,1)\geq 0}\} \qquad D_{\mathbf{V},\mathbf{W}} = \mathbf{diag}\{\mathbb{1}_{(\mathbf{V}^{(0)}+\mathbf{V})(\mathsf{out}_1(x),1)\geq 0}\}$$

which satisfy $\mathsf{out}_1(x) = \mathbf{A}D_{\mathbf{W}}(\mathbf{W}^{(0)} + \mathbf{W})(x, 1)$ and $\mathsf{out}(x) = \mathbf{A}D_{\mathbf{V},\mathbf{W}}(\mathbf{V}^{(0)} + \mathbf{V})(\mathsf{out}_1(x), 1)$.

The proof of Theorem 1 can be divided into three simple steps with parameter choices in Table 1.

---

In this paper, we assume $0 < \alpha \leq \widetilde{O}(\frac{1}{kp_{\mathcal{G}}\mathfrak{C}_{\mathfrak{s}}(\mathcal{G})})$ and choose parameters

$$\sigma_w \in [m^{-1/2+0.01}, m^{-0.01}] \qquad \sigma_v \in [\mathsf{polylog}(m), m^{3/8-0.01}]$$
$$\tau_w \stackrel{\text{def}}{=} \widetilde{\Theta}(kp_{\mathcal{F}}\mathfrak{C}_{\mathfrak{s}}(\mathcal{F})) \geq 1 \qquad \tau_v \stackrel{\text{def}}{=} \widetilde{\Theta}(\alpha kp_{\mathcal{G}}\mathfrak{C}_{\mathfrak{s}}(\mathcal{G})) \leq \frac{1}{\mathsf{polylog}(m)}$$

and they satisfy $\tau_w \in [m^{1/8+0.001}\sigma_w, m^{1/8-0.001}\sigma_w^{1/4}] \qquad \tau_v \in [\sigma_v \cdot (k/m)^{3/8}, \frac{\sigma_v}{\mathsf{polylog}(m)}]$

---

Table 1: Three-layer ResNet parameter choices.
$\sigma_w, \sigma_v$: recall entries of $\mathbf{W}^{(0)}$ and $\mathbf{V}^{(0)}$ are from $\mathcal{N}(0, \sigma_w^2)$ and $\mathcal{N}(0, \sigma_v^2/m)$.
$\tau_w, \tau_v$: the proofs work with respect to $\|\mathbf{W}\|_2 \leq \tau_w$ and $\|\mathbf{V}\|_2 \leq \tau_v$.

In the first step, we prove that for all weight matrices not very far from random initialization (namely, all $\|\mathbf{W}\|_2 \leq \tau_w$ and $\|\mathbf{V}\|_2 \leq \tau_v$), many good "coupling properties" occur. This includes upper bounds on the number of sign changes (i.e., on $\|D_{\mathbf{W}^{(0)}} - D_{\mathbf{W}}\|_0$ and $\|D_{\mathbf{V}^{(0)},\mathbf{W}} - D_{\mathbf{V},\mathbf{W}}\|_0$) as well as vanishing properties such as $\mathbf{A}D_{\mathbf{W}}\mathbf{W}^{(0)}, \mathbf{A}D_{\mathbf{V},\mathbf{W}}\mathbf{V}^{(0)}$ being negligible. We prove such properties using techniques from prior works [4, 6]. Details are in Section C.1.

In the second step, we prove the existence of $\mathbf{W}^*, \mathbf{V}^*$ with $\|\mathbf{W}^*\|_F \leq \frac{\tau_w}{10}$ and $\|\mathbf{V}^*\|_F \leq \frac{\tau_v}{10}$ satisfying $\mathbf{A}D_{\mathbf{W}^{(0)}}\mathbf{W}^*(x, 1) \approx \mathcal{F}(x)$ and $\mathbf{A}D_{\mathbf{V}^{(0)},\mathbf{W}}\mathbf{V}^*(\mathsf{out}_1(x), 1) \approx \alpha\mathcal{G}(\mathsf{out}_1(x))$. This existential

proof relies on an "indicator to function" lemma from [4]; for the purpose of this paper we have to revise it to include a trainable bias term (or equivalently, to support vectors of the form $(x, 1)$). Combining it with the aforementioned vanishing properties, we derive (details are in Section C.2):

$$\mathbf{A}D_{\mathbf{W}}\mathbf{W}^*(x, 1) \approx \mathcal{F}(x) \quad \text{and} \quad \mathbf{A}D_{\mathbf{V},\mathbf{W}}\mathbf{V}^*(\text{out}_1(x), 1) \approx \alpha \mathcal{G}(\text{out}_1(x)) \quad . \tag{6.3}$$

In the third step, consider iteration $t$ of SGD with sample $(x_t, y_t) \sim \mathcal{D}$. For simplicity we assume $\mathsf{OPT} = 0$ so $y_t = \mathcal{H}(x_t)$. One can carefully write down gradient formula, and plug in (6.3) to derive

$$\Xi_t \stackrel{\text{def}}{=} \langle \nabla_{\mathbf{W},\mathbf{V}} \mathsf{Obj}(\mathbf{W}_t, \mathbf{V}_t; (x_t, y_t)), (\mathbf{W}_t - \mathbf{W}^*, \mathbf{V}_t - \mathbf{V}^*)) \rangle$$
$$\geq \tfrac{1}{2}\|\mathcal{H}(x_t) - \text{out}(\mathbf{W}_t, \mathbf{V}_t; x_t)\|_2^2 - 2\|Err_t\|_2^2$$

with $\mathbb{E}\left[\|Err_t\|_2^2\right] \leq \widetilde{\Theta}\left(\alpha^4(kp_{\mathcal{G}}\mathfrak{C}_{\mathfrak{s}}(\mathcal{G}))^4\right)$. This quantity $\Xi_t$ is quite famous in classical mirror descent analysis: for appropriately chosen learning rates, $\Xi_t$ must converge to zero.[14] In other words, by concentration, SGD is capable of finding solutions $\mathbf{W}_t, \mathbf{V}_t$ so that the population risk $\|\mathcal{H}(x_t) - \text{out}(\mathbf{W}_t, \mathbf{V}_t; x_t)\|_2^2$ is as small as $\mathbb{E}[\|Err_t\|_2^2]$. This is why we can obtain population risk $\widetilde{\Theta}\left(\alpha^4(kp_{\mathcal{G}}\mathfrak{C}_{\mathfrak{s}}(\mathcal{G}))^4\right)$ in (6.2). Details are in Section C.3 and C.4.

# 7 Overview of Theorem 2 and 3

We construct the following hard instance. The input $x \in \left\{\frac{\pm 1}{\sqrt{d}}\right\}^d$ in on the (scaled) Boolean cube, and is drawn from distribution $x \sim \mathcal{D} \stackrel{\text{def}}{=} U\left(\left\{\frac{\pm 1}{\sqrt{d}}\right\}^{d_1}\right) \times \mathcal{D}_2$. That is, the first $d_1$ coordinates are drawn uniformly at random from $\{\pm 1/\sqrt{d}\}^{d_1}$, and the last $d - d_1$ coordinates are drawn from an arbitrary distribution $\mathcal{D}_2$. Our hard instance works for a wide range of $d_1$, including for example $d_1 = d$ (uniform distribution over boolean cube) and $d_1 = o(d)$ (only a small subset of the coordinates are uniform). We consider $\mathcal{X} = \{x^{(1)}, \ldots, x^{(N)}\}$ being $N$ i.i.d. samples from $\mathcal{D}$.

Consider the class of target functions $\mathcal{H}(x) = \mathcal{F}(x) + \alpha \mathcal{G}(\mathcal{F}(x))$, where

$$\mathcal{F}(x) = \mathbf{W}^* x \quad and \quad \mathcal{G}(y) = \left(\prod_{j \in [k]} y_j\right)_{i \in [k]} \tag{7.1}$$

where $\mathbf{W}^* = \sqrt{d}(\mathbf{e}_{i_1}, \mathbf{e}_{i_2}, \cdots \mathbf{e}_{i_k})$ for $i_1, i_2, \ldots, i_k \in [d_1]$ are distinct indices chosen from the first $d_1$ coordinates. There are clearly $\binom{d_1}{k}$ many target functions in this class.

Intuitively, $\mathbf{e}_1, \cdots, \mathbf{e}_{d_1}$ represent the directions where the signal possibly lies, where usually the inputs would have high variance; and $\mathbf{e}_{d_1+1}, \ldots, \mathbf{e}_d$ represent the directions that can be view as "background noise", where the distribution can be arbitrary. For example when $d_1 \leq d/2$, such distribution $\mathcal{D}$ can be very different from *Gaussian distribution* or uniform distribution over Boolean cube, yet kernel methods still suffer from high population risk when learning over these distributions comparing to using neural networks.

We first state the population risk for the three-layer ResNet to learn this concept class: Our Theorem 1 implies the following complexity on learning this concept class (after verifying that $\mathfrak{C}_{\mathfrak{s}}(\mathcal{F}) = O(\sqrt{d})$, $p_{\mathcal{F}} = 1$, $\mathfrak{C}_{\mathfrak{s}}(\mathcal{G}) = 2^{O(k)}$, $p_{\mathcal{G}} = 2^k$, see Section D.4).

**Corollary 7.1.** *For every $d \geq d_1 \geq k \geq 2$, for every $\alpha \in \left(0, \frac{1}{\widetilde{\Theta}(2^{O(k)})}\right)$, there exist $M = \mathsf{poly}(d, 2^k, \alpha^{-1})$ satisfying that for every $m \geq M$, for every target functions $\mathcal{H}(x)$ in the class (7.1), with probability at least 0.99 over $\mathbf{A}, \mathbf{W}^{(0)}, \mathbf{V}^{(0)}$ and $\mathcal{X}$, given labels $y^{(n)} = \mathcal{H}(x^{(n)})$ for $n \in [N]$,*

---

[14] Indeed, one can show $\sum_{t=0}^{T-1} \Xi_t \leq O(\eta_w + \eta_v) \cdot T + \frac{\|\mathbf{W}^*\|_F^2}{\eta_w} + \frac{\|\mathbf{V}^*\|_F^2}{\eta_v}$, and thus the right hand side can be made $O(\sqrt{T})$ ignoring other factors.

SGD finds a network $\mathsf{out}(x)$ with population risk

$$\mathbb{E}_{x \sim \mathcal{D}} \|\mathcal{H}(x) - \mathsf{out}(x)\|_2^2 \leq \widetilde{O}(\alpha^4 2^{O(k)}) \quad \text{using } N = \widetilde{\Theta}\left(\frac{k^2 d}{\alpha^8}\right) \text{ samples}.$$

## 7.1 Kernel Method

We restate Theorem 2 as follows.

**Theorem 2** (restated). *For every integers $k, d_1, d, N$ satisfying $2 \leq k \leq d_1 \leq d$ and $N \leq \frac{1}{1000}\binom{d_1}{k}$, for every $\alpha \in (0,1)$, for every $\mathcal{X}$, for every (Mercer) kernels $K_1, \ldots, K_k \colon \mathbb{R}^{d \times d} \to \mathbb{R}$, the following holds for at least 99% of the target functions $\mathcal{H}(x)$ in the class (7.1). For all kernel regression functions*

$$\mathfrak{K}_i(x) = \sum_{n \in [N]} K_i(x, x^{(n)}) \cdot w_{i,n} \quad \text{for } i \in [k],$$

*where weights $w_{i,n} \in \mathbb{R}$ can depend on $\alpha, \mathcal{X}, K$ and the training labels $\{y^{(1)}, \cdots, y^{(N)}\}$, it must suffer from population risk*

$$\mathbb{E}_{x \sim \mathcal{D}} \|\mathcal{H}(x) - \mathfrak{K}(x)\|_2^2 > \alpha^2/16.$$

As an example, when $k \geq 2$ is constant, $d = \Theta(d_1)$ is sufficiently large, and $\alpha = \Theta(d^{-0.1})$,

- Corollary 7.1 says that ResNet achieves regression error $\alpha^{3.9}$ on the true distribution, with $N_{\mathsf{res}} = \widetilde{O}(d^{1.8})$ samples to learn any function in (7.1);
- Theorem 2 says that kernel methods cannot achieve $\alpha^2/16$ error even with $N \leq (N_{\mathsf{res}})^{k/2} \ll o(d^k)$ samples. Hence, to achieve generalization $\alpha^2/16 \gg \alpha^{3.9}$, the sample complexity of any kernel method is at least $N \geq (N_{\mathsf{res}})^{k/2} \gg N_{\mathsf{res}}$.

**Proof Overview.** Our proof of Theorem 2 is relatively simple, and we illustrate the main idea in the case of $d = d_1$. At a high level, given $N \ll \binom{n}{d}$ samples, the kernel regression function only has $N$-degrees of freedom (each with respect to a sample point). Now, since there are possibly $\binom{n}{d}$ many target functions, if the kernel regression learns most of these target functions to some sufficient accuracy, then by some rank counting argument, the degree of freedom is not enough.

## 7.2 Linear Regression Over Feature Mappings

We restate Theorem 3 as follows.

**Theorem 3** (restated). *For every integers $k, d_1, d, D$ satisfying $2 \leq k \leq d_1 \leq d$ and $D \leq \frac{1}{1000}\binom{d_1}{k}$, for every $\alpha \in (0,1)$, for every feature mapping $\phi \colon \mathbb{R}^d \to \mathbb{R}^D$, the following holds for at least 99% of the target functions $\mathcal{H}(x)$ in the class (7.1). For all linear regression functions*

$$\mathfrak{F}_j(x) = w_j^\top \phi(x) \quad \text{for } j \in [k],$$

*where weights $w_j \in \mathbb{R}^D$ can depend on $\alpha$ and $\phi$, it must suffer from population risk*

$$\mathbb{E}_{x \sim \mathcal{D}} \|\mathcal{H}(x) - \mathfrak{F}(x)\|_2^2 > \alpha^2/16.$$

As an example, there exists sufficiently large constant $c > 1$ such that, for every $k \geq 4c$, for every $d_1 \geq d/2$, for every $d \geq \Omega(2^k)$, there exists choice $\alpha = 2^{-\Theta(k)} \cdot d^{-0.001}$ such that

- Corollary 7.1 says that ResNet achieves regression error $\widetilde{O}(\alpha^4 2^{O(k)}) \leq \alpha^{3.9}$ in time $T_{\mathsf{res}} = \mathsf{poly}(d, 2^k, \alpha^{-1}) \leq d^c$ to learn any function in (7.1);

- Theorem 3 says that linear regression over feature mapping cannot achieve regression error $\alpha^2/16$ even if $D = \Omega\left(\binom{d_1}{k}\right) \geq d^{2c}$.

In particular, this means linear regression over feature mappings cannot achieve regression error $\alpha^2/16$ even if $D = (T_{\sf res})^2$. Since a linear regression over $R^D$ normally takes at least time/space $D$ to compute/store, this implies that ResNet is also more time/space efficient than linear regression over feature mappings as well.

Theorem 3 can be proved in the same way as Theorem 2, using exactly the same hard instance, since $\mathfrak{F}(x)$ has exactly $D$-degrees of freedom.

# 8 Experiments

## 8.1 ResNet vs. Kernel Methods vs. Fully-Connected Networks

Consider synthetic data where the feature vectors $x \in \{-1,1\}^{30}$ that are uniformly sampled at random, and labels are generated from a target function $\mathcal{H}(x) = \mathcal{F}(x) + \alpha \mathcal{G}(\mathcal{F}(x)) \in \mathbb{R}^{15}$ satisfying $\mathcal{F}(x) = (x_1 x_2, \ldots, x_{29} x_{30})$ and $\mathcal{G}_i(y) = (-1)^i y_1 y_2 y_3 y_4$ for all $i = 1, 2, \ldots, 15$. In other words, $\mathcal{F}$ is a degree-2 parity function over 30 dimensions, and $\mathcal{G}$ is a degree-4 parity function over 15 dimensions.

**Neural Networks Algorithms.** Recall in our positive result on three-layer ResNet (see Theorem 1 and Footnote 12), to prove the strongest result, we only train hidden weights $\mathbf{W}$ and $\mathbf{V}$ but not the output layer $\mathbf{A}$. One can naturally extend this to show that Theorem 1 also holds when $\mathbf{W}, \mathbf{V}, \mathbf{A}$ are jointly trained. For such reason, we implement both algorithms: `3resnet(hidden)` for training only $\mathbf{W}, \mathbf{V}$ and `3resnet(all)` for training all $\mathbf{W}, \mathbf{V}, \mathbf{A}$. This is similar for two-layer and three-layer fully-connected networks, where previously the strongest theoretical work is in terms of training only hidden weights [4], so we implement both (`all`) and (`hidden`) for them.

**Kernel Methods.** We implement *conjugate kernel*, which corresponds to training only the last (output) layer [12]; as well as neural tangent kernel (NTK), in which we train all the layers [21].

**Setup.** We choose the network width (i.e., parameter $m$) in the range $m \in \{20, 50, 100, 200, \ldots\}$ until the largest possible value $m$ that fits into a 16GB GPU memory. We choose the popular random initialization: entries of $\mathbf{A}, \mathbf{V}, \mathbf{W}$ (and their corresponding bias terms) are all i.i.d. from $\mathcal{N}(1, \frac{1}{m})$.[15] We use similar initializations for two and three-layer networks.

We use the default SGD optimizer of pytorch, with momentum 0.9, mini-batch size 50. We carefully run each algorithm with respect to learning rates and weight decay parameters in the set $\{10^{-k}, 2 \cdot 10^{-k}, 5 \cdot 10^{-k} : k \in \mathbb{Z}\}$, and present the best one in terms of testing accuracy. In each parameter setting, we run SGD for 800 epochs, and decrease the learning rate by 10 on epoch 400.

**Experiment 1: Performance Comparison.** Since it is unfair to compare neural network training "with respect to hidden weights only" vs. "with respect to all weights", we conduct two experiments. The first experiment is on training all layers vs. kernel methods, see Figure 2(a); and the second experiment is on training hidden layers vs. kernel methods, see Figure 2(b). We use $N = 500$ training samples for the former case and $N = 1000$ samples for the latter case, because training the last layer together gives more power to a neural network.

In both experiments, we choose $\alpha = 0.3$ and $k = 15$ so that test error $k\alpha^2 = 1.35$ is a threshold for detecting whether the trained model has successfully learned $\alpha \mathcal{G}(\mathcal{F}(x))$ or not. If the model has not learned $\alpha \mathcal{G}(\mathcal{F}(x))$ to any non-trivial accuracy, then the error is $\alpha$ per output coordinate, totaling to $k\alpha^2$ in regression error.

---

[15]This corresponds to choosing the standard deviation as $\frac{1}{\sqrt{\texttt{fan\_in}} + \sqrt{\texttt{fan\_out}}}$. Some practitioners also use $\frac{1}{\sqrt{\texttt{fan\_in}}}$ as the standard deviation. We have included an experiment with respect to that choice in our V1/V2 of this paper.

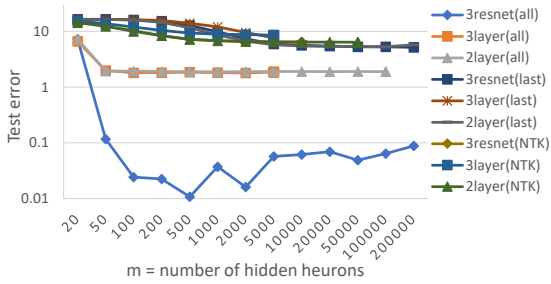
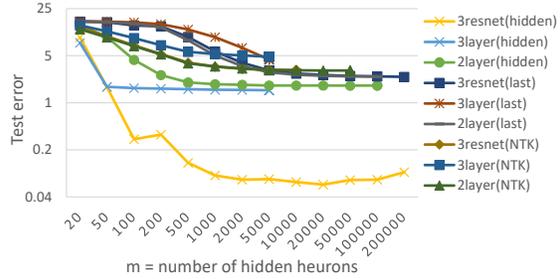

(a) $N = 500$, train all layers vs. kernel methods

(b) $N = 1000$, train hidden layers vs. kernel methods

Figure 2: Performance comparison. `3resnet` stands for our three-layer ResNet and `3layer`/`2layer` stands for three and two-layer fully connected networks. `(all)` stands for training all layers, `(hidden)` stands for training only hidden layers, `(last)` stands for training only the last output layer, and `(NTK)` stands for training all layers in the *neural tangent kernel* [21]. We emphasize that `(last)` is a kernel method and corresponds to the *conjugate kernel* [12]. Experiment setup is in Section 8.1.

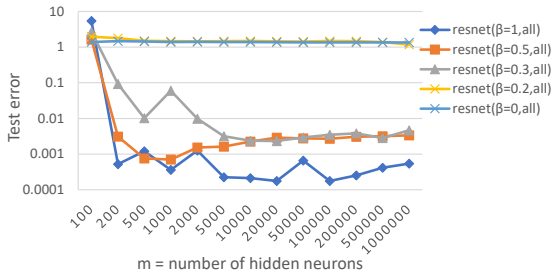
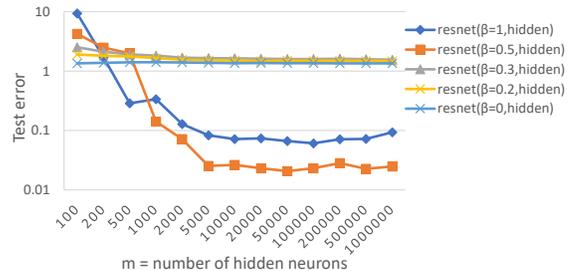

(a) training all layers of `3resnet`

(b) training hidden layers of `3resnet`

Figure 3: Sensitivity test on $\alpha$. Using the same choice of $\mathcal{F}(x)$ and $\mathcal{G}(y)$ from Section 8.1, we choose target function $\mathcal{H}(x) = \beta\mathcal{F}(x) + \alpha\mathcal{G}(\mathcal{F}(x))$ with $\alpha = 0.3$ and varying $\beta \in [0, 1]$.

From Figure 2, it is clear that for our choice of $N$, training a three-layer ResNet is the only method among the ones we compare that can learn $\alpha\mathcal{G}(\mathcal{F}(x))$ (even only non-trivially). All kernel methods fall far behind even when the network width $m$ is large.

**Experiment 2: Sensitivity on $\alpha$.** One key assumption of this paper is to have $\alpha$ to be sufficiently small, so that ResNet can perform hierarchical learning, by first learning the base signal $\mathcal{F}$, which is simpler and contributes more to the target, and then learning the composite signal $\alpha\mathcal{G}(\mathcal{F})$, which is more complicated but contributes less.

In Figure 3, we verify that this assumption is indeed necessary. Instead of varying $\alpha$ (which will change the error magnitude), we define $\mathcal{H}(x) = \beta\mathcal{F}(x) + \alpha\mathcal{G}(\mathcal{F}(x))$ and let $\beta$ vary between 0 and 1. As shown in Figure 3, when $\alpha \lesssim \beta$, the base signal is larger than the composite signal, so indeed ResNet can perform hierachical learning; in contrast, when $\alpha \gtrsim \beta$, learning the composite signal becomes practically impossible.

**Other Findings.** Although this paper proves theoretical separation between three-layer ResNet and kernel methods (and it is verified by Figure 2), we do not yet have

- theoretical separation between two/three-layer fully-connected networks and kernel methods;

- theoretical separation between three-layer ResNet and two/three-layer networks.

It seems in practice such separations do exist (as observed in Figure 2). We leave these as future research directions.

## 8.2 SGD Does Not Converge To Minimal Norm Solutions

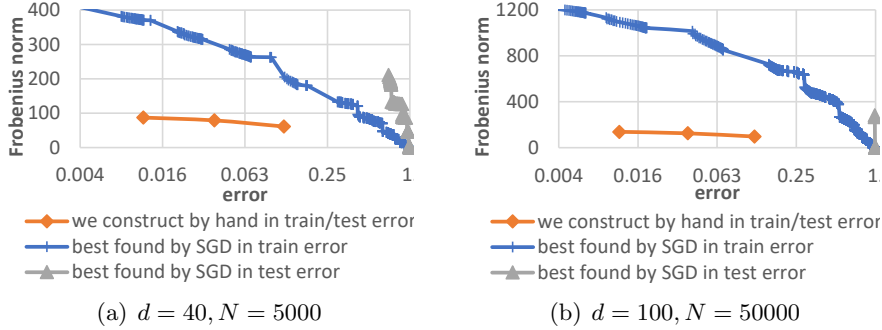

(a) $d = 40, N = 5000$      (b) $d = 100, N = 50000$

Figure 4: SGD cannot find solutions with Frobenius norm comparable to what we construct by hand.

We give a simple experiment to show that optimization methods (such as SGD) do not necessarily converge to minimal complexity solutions.

Consider two-layer neural networks $F(W; x) = a^\top \sigma(Wx)$ where $W \in \mathbb{R}^{m \times d}$ and $a \in \{\frac{\pm 1}{\sqrt{m}}\}^m$ is an arbitrary vector with exactly $m/2$ positive and $m/2$ negative values. For simplicity, we focus on the case when $x$ is of norm 1 and we only train $W$ keeping $a$ fixed.

Consider a simple data distribution where each $x_i$ is independently drawn from $\{\frac{\pm 1}{\sqrt{d}}\}$ for some $d \geq 6$. Consider labels $y \in \{-1, +1\}$ being generated from some target function $y = \mathcal{F}(x) \stackrel{\text{def}}{=} d^3 x_{i_1} x_{i_2} x_{i_3} x_{i_4} x_{i_5} x_{i_6}$ for some distinct indices $i_1, i_2, i_3, i_4, i_5, i_6 \in [d]$.

It is a simple experimental exercise to verify that, for every even $m \geq 200$ and every $d \geq 6$, there exist[16]

- $W^* \in \mathbb{R}^{m \times d}$ with $\|W^*\|_F \approx 9.7\sqrt{d}$ satisfying $\mathbb{E}_{(x,y)} \left[ |F(W; x) - y|^2 \right] \leq 0.12$.
- $W^* \in \mathbb{R}^{m \times d}$ with $\|W^*\|_F \approx 12.5\sqrt{d}$ satisfying $\mathbb{E}_{(x,y)} \left[ |F(W; x) - y|^2 \right] \leq 0.037$.
- $W^* \in \mathbb{R}^{m \times d}$ with $\|W^*\|_F \approx 13.8\sqrt{d}$ satisfying $\mathbb{E}_{(x,y)} \left[ |F(W; x) - y|^2 \right] \leq 0.011$.

Using simple Rademacher complexity argument, the above existential statement implies if we focus only on matrices $W$ with $\|W\|_F \leq 9.7\sqrt{d}$, then given $N$ training samples the Rademacher complexity is at most $\frac{\frac{2}{\sqrt{m}} \sum_{j \in [m]} \|W_j\|_2}{\sqrt{N}} \leq \frac{2\|W\|_F}{\sqrt{N}}$.[17] This implies, for any $m \geq 200$ and $d \geq 6$, if $N = O(d)$ samples are given and if SGD finds any close-to-minimal complexity solution (i.e. with F-norm within some constant times $\sqrt{d}$) that performs well on the training set, then it also generalizes to give small test error (i.e. test error $< 0.3$).

Unfortunately, one can experimentally verify that, even for $d = 40$ and $N = 5000$, starting from random initialization, even after searching learning rates and weight decay parameters in the set $\{10^{-k}, 2 \cdot 10^{-k}, 5 \cdot 10^{-k} : k \in \mathbb{Z}\}$, searching network size $m$ in $\{200, 500, 1000, 2000, \ldots, 100000\}$:

---

[16]This can be done by first considering $m = 200$ and $d = 6$. Experimentally one can easily use SGD to train such two-layer networks to obtain some $W^*$ with such test errors. Then, for general $d > 6$, one can pad $W^*$ with $d - 6$ zero columns; and for general $m > 200$, one can duplicate the rows of $W^*$ and re-scale.

[17]This can found for instance in [18, 32]. A cleaner one page proof can be found in the lecture notes [30].

- SGD cannot find solution with test error better than 0.69 (see Figure 4(a)), and
- SGD cannot find solution with small training error and small Frobenius norm (see Figure 4(a)). Thus, SGD starting from random initialization *fails* to find the minimal complexity solution.

We also tried $d = 100$ and $N = 50000$ (where $N$ is the same comparing to the standard CIFAR10/100 datasets), and this time we choose mini-batch 100 to speed up training. Even after searching learning rates and weight decay parameters in the set $\{10^{-k}, 2 \cdot 10^{-k}, 5 \cdot 10^{-k} : k \in \mathbb{Z}\}$, searching network size $m$ in $\{200, 500, 1000, 2000, \dots, 50000\}$:

- SGD cannot find solution with test error better than 0.98 (see Figure 4(b)), and
- SGD cannot find solution with small training error and small Frobenius norm (see Figure 4(b)).

# Appendix: Complete Proofs

In Appendix A we give some more information about our concept class and complexity measure.
In Appendix B we review some simple lemmas from probability theory.
In Appendix C we give our full proof to Theorem 1.
In Appendix D we give our full proof to Theorem 2.
In Appendix E we include a variant of the existential lemma from prior work, and include its proof only for completeness' sake.

## A  Complexity and Concept Class

In this section we introduce an alternative (but bigger) concept class.

**Definition A.1.** *We say $\mathcal{F} \colon \mathbb{R}^d \to \mathbb{R}^k$ has general complexity $(p, \mathfrak{C}_\mathfrak{s}(\mathcal{F}), \mathfrak{C}_\varepsilon(\mathcal{F}))$ if for each $r \in [k]$,*

$$\mathcal{F}_r(x) = \sum_{i=1}^p a_{r,i}^* \cdot \mathcal{F}_{r,i}\left(\frac{\langle w_{1,i}^*, (x,1)\rangle}{\|(x,1)\|_2}\right) \cdot \langle w_{2,i}^*, (x,1)\rangle \ ,$$

*where each $a_{r,i}^* \in [-1, 1]$, each $w_{1,i}^*, w_{2,i}^* \in \mathbb{R}^{d+1}$ has Euclidean norm 1, each $\mathcal{F}_{r,i} \colon \mathbb{R} \to \mathbb{R}$ is a smooth function with only zero-order and odd-order terms in its Taylor expansion at point zero, and $\mathfrak{C}_\varepsilon(\mathcal{F}) = \max_{r,i}\{\mathfrak{C}_\varepsilon(\mathcal{F}_{r,i})\}$ and $\mathfrak{C}_\mathfrak{s}(\mathcal{F}) = \max_{r,i}\{\mathfrak{C}_\mathfrak{s}(\mathcal{F}_{r,i})\}$.*

**Concept 2.** *$\mathcal{H}$ is given by two smooth functions $\mathcal{F}, \mathcal{G} : \mathbb{R}^k \to \mathbb{R}^k$ and a value $\alpha \in \mathbb{R}_+$:*

$$\mathcal{H}(x) = \mathcal{F}(x) + \alpha \mathcal{G}(\mathcal{F}(x)) \ , \tag{A.1}$$

*where where $\mathcal{F}$ and $\mathcal{G}$ respectively have general complexity $(p_\mathcal{F}, \mathfrak{C}_\mathfrak{s}(\mathcal{F}), \mathfrak{C}_\varepsilon(\mathcal{G}))$ and $(p_\mathcal{G}, \mathfrak{C}_\mathfrak{s}(\mathcal{G}), \mathfrak{C}_\varepsilon(\mathcal{G}))$. We further assume $\|\mathcal{F}(x)\|_2 \leq \mathfrak{B}_\mathcal{F}$ for all $(x, y) \sim \mathcal{D}$.*

We have the following lemma which states that Concept 1 is a special case of Concept 2 (with constant factor 2 blow up).

**Lemma A.2.** *Under Concept 1, we can construct $\mathcal{F}', \mathcal{G}'$ satisfying Concept 2 with general complexity $(2p_\mathcal{F}, \mathfrak{C}_\mathfrak{s}(\mathcal{F}), \mathfrak{C}_\varepsilon(\mathcal{G}))$ and $(2p_\mathcal{G}, \mathfrak{C}_\mathfrak{s}(\mathcal{G}), \mathfrak{C}_\varepsilon(\mathcal{G}))$ and with $\mathfrak{B}_\mathcal{F} = 1$.*

*Proof of Lemma A.2.* Lemma A.2 is a simple corollary of the following claim.

Given any $\mathcal{F}\colon \mathbb{R}^d \to \mathbb{R}^k$ where for each $r \in [k]$:

$$\mathcal{F}_r(x) = \sum_{i=1}^{p} a_{r,i}^* \cdot \mathcal{F}_{r,i}\left(\frac{\langle w_i^*, x\rangle}{\sqrt{2}}\right) \ ,$$

where each $a_{r,i}^* \in [-1, 1]$, each $w_i^* \in \mathbb{R}^d$ has Euclidean norm 1, each $\mathcal{F}_{r,i}\colon \mathbb{R} \to \mathbb{R}$ is a smooth function. Then, there exists some $\mathcal{F}'\colon \mathbb{R}^d \to \mathbb{R}^k$ such that:

- $\mathcal{F}(x) = \mathcal{F}'(x)$ for all unit vectors $x \in \mathbb{R}^d$; and
- $\mathcal{F}'$ has general complexity $(2p, \mathfrak{C}_\mathfrak{s}(\mathcal{F}), \mathfrak{C}_\varepsilon(\mathcal{F}))$ where $\mathfrak{C}_\varepsilon(\mathcal{F}) = \max_{r,i}\{\mathfrak{C}_\varepsilon(\mathcal{F}_{r,i})\}$ and $\mathfrak{C}_\mathfrak{s}(\mathcal{F}) = \max_{r,i}\{\mathfrak{C}_\mathfrak{s}(\mathcal{F}_{r,i})\}$.

Below we prove that the above claim holds. For each $\mathcal{F}_{r,i}(\cdot)$ suppose we have $\mathcal{F}_{r,i}(z) = \sum_{i=0}^{\infty} c_i z^i$ as its Taylor expansion, then we can write

$$\mathcal{F}_{r,i}(z) = \mathcal{F}_{r,i}^+(z) + z \cdot \mathcal{F}_{r,i}^-(z) \stackrel{\text{def}}{=} \left(\sum_{i=0,1,3,5,7,\ldots} c_i z^i\right) + z \cdot \left(\sum_{i=2,4,6,8,\ldots} c_i z^{i-1}\right) \ .$$

From this expansion we see that both $\mathcal{F}_{r,i}^+$ and $\mathcal{F}_{r,i}^-$ have only zero-order or odd-order terms in its Taylor expansion at zero. We can define $\mathcal{F}'\colon \mathbb{R}^d \to \mathbb{R}^k$ where

$$\mathcal{F}'_r(x) = \sum_{i=1}^{p} a_{r,i}^* \cdot \left(\mathcal{F}_{r,i}^+\left(\frac{\langle (w_i^*, 0), (x, 1)\rangle}{\|(x,1)\|_2}\right) \cdot \langle (\vec{0}, 1), (x, 1)\rangle + \frac{1}{\sqrt{2}}\mathcal{F}_{r,i}^-\left(\frac{\langle (w_i^*, 0), (x, 1)\rangle}{\|(x,1)\|_2}\right) \cdot \langle (w_i^*, 0), (x, 1)\rangle\right)$$

It is a simple exercise to verify that $\mathcal{F}'(x) = \mathcal{F}(x)$ for all unit vectors $x$. □

We also state some simple properties regarding our complexity measure.

**Fact A.3.** *If $\mathcal{F}\colon \mathbb{R}^d \to \mathbb{R}^k$ has general complexity $(p, \mathfrak{C}_\mathfrak{s}(\mathcal{F}), \mathfrak{C}_\varepsilon(\mathcal{F}))$, then for every $x, y \in \mathbb{R}^d$, it satisfies $\|\mathcal{F}(x)\|_2 \leq \sqrt{k}p\mathfrak{C}_\mathfrak{s}(\mathcal{F}) \cdot \|x\|_2$ and $\|\mathcal{F}(x) - \mathcal{F}(y)\|_2 \leq \sqrt{k}p\mathfrak{C}_\mathfrak{s}(\mathcal{F}) \cdot \|x - y\|_2$.*

*Proof of Fact A.3.* The boundedness of $\|\mathcal{F}(x)\|_2$ is trivial so we only focus on $\|\mathcal{F}(x) - \mathcal{F}(y)\|_2$. For each component $g(x) = \mathcal{F}_{r,i}\left(\frac{\langle w_{1,i}^*, (x,1)\rangle}{\|(x,1)\|_2}\right) \cdot \langle w_{2,i}^*, (x, 1)\rangle$, denoting by $w_1^*$ as the first $d$ coordinate of $w_{1,i}^*$, and by $w_{2,i}^*$ as the first $d$ coordinates of $w_{2,i}^*$, we have

$$g'(x) = \mathcal{F}_{r,i}\left(\frac{\langle w_{1,i}^*, (x,1)\rangle}{\|(x,1)\|_2}\right) \cdot w_2^*$$
$$+ \langle w_{2,i}^*, (x, 1)\rangle \cdot \mathcal{F}'_{r,i}\left(\frac{\langle w_{1,i}^*, (x,1)\rangle}{\|(x,1)\|_2}\right) \cdot \frac{w_1^* \cdot \|(x,1)\|_2 - \langle w_{1,i}^*, (x,1)\rangle \cdot (x,1)/\|(x,1)\|_2}{\|(x,1)\|_2^2}$$

This implies

$$\|g'(x)\|_2 \leq \left|\mathcal{F}_{r,i}\left(\frac{\langle w_{1,i}^*, (x,1)\rangle}{\|(x,1)\|_2}\right)\right| + 2\left|\mathcal{F}'_{r,i}\left(\frac{\langle w_{1,i}^*, (x,1)\rangle}{\|(x,1)\|_2}\right)\right| \leq 3\mathfrak{C}_\mathfrak{s}(\mathcal{F}_{r,i}) \ .$$

As a result, $|\mathcal{F}_r(x) - \mathcal{F}_r(y)| \leq 3p\mathfrak{C}_\mathfrak{s}(\mathcal{F}_{r,i})$. □

# B  Probability Theory Review

The following concentration of chi-square distribution is standard.

**Proposition B.1** (chi-square concentration). *If $g \sim \mathcal{N}(0, \mathbf{I})$ is $m$-dimensional, then for every $t \geq 1$*
$$\mathbf{Pr}[\|g\|_2^2 - m \geq 2\sqrt{mt} + 2t] \leq e^{-t}$$

The following norm bound on random Gaussian matrix is standard.

**Proposition B.2.** *If $\mathbf{M} \in \mathbb{R}^{n \times m}$ is a random matrix where $\mathbf{M}_{i,j}$ are i.i.d. from $\mathcal{N}(0,1)$. Then,*

- *For any $t \geq 1$, with probability $\geq 1 - e^{-\Omega(t^2)}$ it satisfies $\|\mathbf{M}\|_2 \leq O(\sqrt{n} + \sqrt{m}) + t$.*
- *If $1 \leq s \leq O\big(\frac{m}{\log^2 m}\big)$, then with probability $\geq 1 - e^{-\Omega(n + s \log^2 m)}$ it satisfies $\|\mathbf{M}v\|_2 \leq O(\sqrt{n} + \sqrt{s} \log m) \cdot \|v\|_2$ for all $s$-sparse vectors $v \in \mathbb{R}^m$.*

*Proof.* The first statement can be found for instance in [34, Proposition 2.4]. As for the second statement, it suffices for us to consider all $\binom{m}{s}$ possible $n \times s$ sub-matrices of $\mathbf{M}$, each applying the first statement, and then taking a union bound. □

The following concentration is proved for instance in [4].

**Lemma B.3** (Gaussian indicator concentration). *Let $(n_1, \alpha_1, a_{1,1}, a_{2,1}), \cdots, (n_m, \alpha_m, a_{1,m}, a_{2,m})$ be $m$ i.i.d. samples from some distribution, where within a 4-tuples:*
- *the marginal distribution of $a_{1,i}$ and $a_{2,i}$ is standard Gaussian $\mathcal{N}(0,1)$;*
- *$n_i$ and $\alpha_i$ are not necessarily independent;*
- *$a_{1,i}$ and $a_{2,i}$ are independent; and*
- *$n_i$ and $\alpha_i$ are independent of $a_{1,i}$ and $a_{2,i}$.*

*Suppose $h \colon \mathbb{R} \to [-L, L]$ is a fixed function. Then, for every $B \geq 1$:*
$$\mathbf{Pr}\left[\left|\left(\sum_{i \in [m]} a_{1,i} a_{2,i} \mathbb{1}[n_i \geq 0] h(\alpha_i)\right)\right| \geq BL(\sqrt{m} + B)\right] \leq 4e^{-B^2/8}$$
*and*
$$\mathbf{Pr}\left[\left|\left(\sum_{i \in [m]} a_{1,i}^2 \mathbb{1}[n_i \geq 0] h(\alpha_i)\right) - m\,\mathbb{E}[a_{1,1}^2 \mathbb{1}[n_1 \geq 0] h(\alpha_1)]\right| \geq BL(\sqrt{m} + B)\right] \leq 4e^{-B^2/8}.$$

*Proof of Lemma B.3.* Let us consider a fixed $n_1, \alpha_1, \cdots, n_m, \alpha_m$, then since each $|\mathbb{1}[n_i \geq 0]h(\alpha_i)| \leq L$, by Gaussian chaos variables concentration bound (e.g., Example 2.15 in [31]) we have that
$$\mathbf{Pr}\left[\left|\left(\sum_{i \in [m]} a_{1,i} a_{2,i} \mathbb{1}[n_i \geq 0] h(\alpha_i)\right)\right| \geq BL(\sqrt{m} + B)\,\bigg|\,\{n_i, \alpha_i\}_{i \in [m]}\right] \leq 4e^{-B^2/8}.$$
Since this holds for every choice of $\{n_i, \alpha_i\}_{i \in [m]}$ we can complete the proof. The second inequality follows from sub-exponential concentration bounds. □

The next proposition at least traces back to [5] and was stated for instance in [6].

**Proposition B.4.** *Suppose $\delta \in [0,1]$ and $g^{(0)} \in \mathbb{R}^m$ is a random vector $g^{(0)} \sim \mathcal{N}(0, \frac{\mathbf{I}}{m})$. With probability at least $1 - e^{-\Omega(m\delta^{2/3})}$, for all vectors $g' \in \mathbb{R}^m$ with $\|g'\|_2 \leq \delta$, letting $D' \in \mathbb{R}^{m \times m}$ be the diagonal matrix where $(D')_{k,k} = \mathbb{1}_{(g^{(0)} + g')_k \geq 0} - \mathbb{1}_{(g^{(0)})_k \geq 0}$ for each $k \in [m]$, we have*
$$\|D'\|_0 \leq O(m\delta^{2/3}) \quad \text{and} \quad \|D'g^{(0)}\|_2 \leq \|g'\|_2 \ .$$

*Proof of Proposition B.4.* Observe that $(D')_{j,j}$ is non-zero for some $j \in [m]$ only if

$$|g'_j| > |(g^{(0)})_j| \ . \tag{B.1}$$

Therefore, denoting by $x = D'g^{(0)}$, for each $j \in [m]$ such that $x_j \neq 0$, we must have $|x_j| = |(g^{(0)})_j| \leq |(g')_j|$ so we have $\|x\|_2 \leq \|g'\|_2$ .

Let $\xi \leq \frac{1}{2\sqrt{m}}$ be a constant parameter to be chosen later.

- We denote by $S_1 \subseteq [m]$ the index sets where $j$ satisfies $|(g^{(0)})_j| \leq \xi$. Since we know $(g^{(0)})_j \sim \mathcal{N}(0, 1/m)$, we have $\mathbf{Pr}[|(g^{(0)})_j| \leq \xi] \leq O(\xi\sqrt{m})$ for each $j \in [m]$. Using Chernoff bound for all $j \in [m]$, we have with probability at least $1 - e^{-\Omega(m^{3/2}\xi)}$,

$$|S_1| = \left|\left\{i \in [m]\colon |(g^{(0)})_j| \leq \xi\right\}\right| \leq O(\xi m^{3/2}) \ .$$

- We denote by $S_2 \subseteq [m] \setminus S_1$ the index set of all $j \in [m] \setminus S_1$ where $(D')_{j,j} \neq 0$. Using (B.1), we have for each $j \in S_2$ it satisfies $|(g')_j| \geq |(g^{(0)})_j| \geq \xi$ . This means $|S_2| \leq \frac{\|g'_1\|_2^2}{\xi^2}$ .

From above, we have $\|D'\|_0 \leq |S_1| + |S_2| \leq O\big(\xi m^{3/2} + \frac{\delta^2}{\xi^2}\big)$. Choosing $\xi = \frac{\delta^{2/3}}{2m^{1/2}}$ gives the desired result. □

## C Theorem 1 Proof Details

In the analysis, let us define a diagonal matrices

$$D_{\mathbf{W}^{(0)}} = \mathbf{diag}\{\mathbb{1}_{\mathbf{W}^{(0)}(x,1)\geq 0}\} \qquad D_{\mathbf{V}^{(0)},\mathbf{W}} = \mathbf{diag}\{\mathbb{1}_{\mathbf{V}^{(0)}(\mathsf{out}_1(x),1)\geq 0}\}$$
$$D_{\mathbf{W}} = \mathbf{diag}\{\mathbb{1}_{(\mathbf{W}^{(0)}+\mathbf{W})(x,1)\geq 0}\} \qquad D_{\mathbf{V},\mathbf{W}} = \mathbf{diag}\{\mathbb{1}_{(\mathbf{V}^{(0)}+\mathbf{V})(\mathsf{out}_1(x),1)\geq 0}\}$$

which satisfy $\mathsf{out}_1(x) = \mathbf{A}D_{\mathbf{W}}(\mathbf{W}^{(0)}+\mathbf{W})(x,1)$ and $\mathsf{out}(x) = \mathbf{A}D_{\mathbf{V},\mathbf{W}}(\mathbf{V}^{(0)}+\mathbf{V})(\mathsf{out}_1(x),1)$.

Throughout the proof, we assume $m \geq \mathsf{poly}(\mathfrak{C}_\alpha(\mathcal{F}), \mathfrak{C}_\alpha(\mathcal{G}), p_\mathcal{G}, p_\mathcal{F}, k, \alpha^{-1})$.

### C.1 Coupling

In this subsection we present our coupling lemma. It shows that for all weight matrices not very far from random initialization (namely, all $\|\mathbf{W}\|_2 \leq \tau_w$ and $\|\mathbf{V}\|_2 \leq \tau_v$), many good properties occur. This includes upper bounds on the number of sign changes (i.e., on $\|D_{\mathbf{W}^{(0)}} - D_{\mathbf{W}}\|_0$ and $\left\|D_{\mathbf{V}^{(0)},\mathbf{W}} - D_{\mathbf{V},\mathbf{W}}\right\|_0$) as well as vanishing properties such as $\mathbf{A}D_{\mathbf{W}}\mathbf{W}^{(0)}, \mathbf{A}D_{\mathbf{V},\mathbf{W}}\mathbf{V}^{(0)}$ being negligible. We prove such properties using techniques from prior works [4, 6].

**Lemma C.1** (Coupling). *Suppose $\tau_w \geq 1$, $\tau_w \in [m^{1/8+0.001}\sigma_w, m^{1/8-0.001}\sigma_w^{1/4}]$, and $\tau_v \in [\sigma_v \cdot (k/m)^{3/8}, \sigma_v]$. Then, for every fixed $x$, with high probability over $\mathbf{A}, \mathbf{W}^{(0)}, \mathbf{V}^{(0)}$, we have that for all $\mathbf{W}, \mathbf{V}$ satisfying $\|\mathbf{W}\|_2 \leq \tau_w$ and $\|\mathbf{V}\|_2 \leq \tau_v$, it holds that*

(a) $\|D_{\mathbf{W}^{(0)}} - D_{\mathbf{W}}\|_0 \leq O((\tau_w/\sigma_w)^{2/3}m^{2/3})$

(b) $\left\|\mathbf{A}D_{\mathbf{W}}\mathbf{W}(x,1) - \mathbf{A}D_{\mathbf{W}}((\mathbf{W}^{(0)}+\mathbf{W})(x,1))\right\|_2 \leq \widetilde{O}\left(\frac{\tau_w(\tau_w/\sigma_w)^{1/3}}{m^{1/6}}\right) \leq O(m^{-0.001})$

(c) $\|\mathsf{out}_1(x)\|_2 = \left\|\mathbf{A}D_{\mathbf{W}}(\mathbf{W}^{(0)}+\mathbf{W})(x,1)\right\|_2 \leq O(\tau_w)$

(d) $\left\|D_{\mathbf{V}^{(0)},\mathbf{W}} - D_{\mathbf{V},\mathbf{W}}\right\|_0 \leq O((\tau_v/\sigma_v)^{2/3}m)$

(e) $\left\|\mathbf{A}D_{\mathbf{V},\mathbf{W}}\mathbf{V}(\mathsf{out}_1(x),1) - \mathbf{A}D_{\mathbf{V},\mathbf{W}}(\mathbf{V}^{(0)}+\mathbf{V})(\mathsf{out}_1(x),1)\right\|_2 \leq \widetilde{O}\left(\tau_v(\tau_v/\sigma_v)^{1/3}\right)\cdot(\|\mathsf{out}_1(x)\|_2+1)$

(f) $\|\mathbf{A}D_{\mathbf{V},\mathbf{W}}\mathbf{V}^{(0)}\|_2 \leq \tilde{O}\left(\tau_v(\tau_v/\sigma_v)^{1/3}\right)$

(g) $\|\mathbf{A}D_{\mathbf{V},\mathbf{W}}(\mathbf{V}^{(0)} + \mathbf{V})(\mathsf{out}_1(x), 1)\|_2 \leq \tilde{O}\left(\tau_v(\|\mathsf{out}_1(x)\|_2 + 1)\right)$

*Proof.*

(a) Using basic probability argument (appropriately scaling and invoking Proposition B.4) we have
$$\|D_{\mathbf{W}} - D_{\mathbf{W}^{(0)}}\|_0 \leq O\left(\left(\frac{\tau_w}{\sigma_w\sqrt{m}}\right)^{2/3} \cdot m\right) = O((\tau_w/\sigma_w)^{2/3} m^{2/3}) \ .$$

(b) We write
$$\mathbf{A}D_{\mathbf{W}}\mathbf{W}(x, 1) - \mathbf{A}D_{\mathbf{W}}(\mathbf{W}^{(0)} + \mathbf{W})(x, 1) = -\mathbf{A}D_{\mathbf{W}^{(0)}}\mathbf{W}^{(0)}(x, 1) + \mathbf{A}(D_{\mathbf{W}^{(0)}} - D_{\mathbf{W}})\mathbf{W}^{(0)}(x, 1)$$
For the first term, we have $\|D_{\mathbf{W}^{(0)}}\mathbf{W}^{(0)}(x, 1)\|_2 \leq \|\mathbf{W}^{(0)}(x, 1)\|_2 \leq O(\sigma_w\sqrt{m})$ with high probability due to concentration of chi-square distribution, and then using the randomness of $\mathbf{A}$ and applying concentration of chi-square distribution again, we have $\|\mathbf{A}D_{\mathbf{W}^{(0)}}\mathbf{W}^{(0)}(x, 1)\|_2 \leq \tilde{O}(\sqrt{k}\sigma_w)$ with high probability.

For the second term, invoking Proposition B.4 again, we have
$$\|(D_{\mathbf{W}} - D_{\mathbf{W}^{(0)}})\mathbf{W}^{(0)}(x, 1)\|_2 \leq \|\mathbf{W}(x, 1)\|_2 \leq \tau_w$$

Recall for every $s$-sparse vectors $y$, it satisfies $\|\mathbf{A}y\|_2 \leq \tilde{O}(\frac{\sqrt{s}}{\sqrt{m}}) \cdot \|y\|_2$ with high probability (see Proposition B.2). This implies
$$\|\mathbf{A}(D_{\mathbf{W}} - D_{\mathbf{W}^{(0)}})\mathbf{W}^{(0)}(x, 1)\|_2 \leq \tilde{O}(\frac{\sqrt{s}}{\sqrt{m}}) \cdot \|\mathbf{W}(x, 1)\|_2 \leq \tilde{O}(\frac{\sqrt{s}}{\sqrt{m}}\tau_w)$$
for $s = O\left(\left(\frac{\tau_w}{\sigma_w\sqrt{m}}\right)^{2/3} \cdot m\right)$. Together, we have
$$\|\mathbf{A}D_{\mathbf{W}}\mathbf{W}(x, 1) - \mathbf{A}D_{\mathbf{W}}((\mathbf{W}^{(0)} + \mathbf{W})(x, 1))\|_2 \leq \tilde{O}\left(\frac{\tau_w(\tau_w/\sigma_w)^{1/3}}{m^{1/6}} + \sqrt{k}\sigma_w\right) \ .$$

(c) We use Lemma C.1b together with $\|\mathbf{A}D_{\mathbf{W}}\mathbf{W}(x, 1)\|_2 \leq \|\mathbf{A}\|_2\|\mathbf{W}\|_2 \leq O(\tau_w)$, where the property $\|\mathbf{A}\|_2 \leq O(1)$ holds with high probability using Proposition B.2.

(d) Recall $D_{\mathbf{V}^{(0)},\mathbf{W}} = \mathbf{diag}\{\mathbb{1}_{\mathbf{V}^{(0)}(\mathsf{out}_1(x),1)\geq 0}\}$ and $D_{\mathbf{V},\mathbf{W}} = \mathbf{diag}\{\mathbb{1}_{(\mathbf{V}^{(0)}+\mathbf{V})(\mathsf{out}_1(x),1)\geq 0}\}$. Let us denote by $z = (\mathsf{out}_1(x), 1)$. We know that if $z \in \mathbb{R}^{k+1}$ is a fixed vector (as opposed to depending on $\mathbf{W}^{(0)}$ and $\mathbf{W}$), then owing to Proposition B.4
$$\|D_{\mathbf{V},\mathbf{W}} - D_{\mathbf{V}^{(0)},\mathbf{W}}\|_0 \leq O\left(\left(\frac{\tau_v}{\sigma_v}\right)^{2/3} \cdot m\right) \tag{C.1}$$
with probability at least $1 - e^{-\Omega(m^{2/3})}$. This means, taking $\varepsilon$-net over all possible unit vectors $z \in \mathbb{R}^{k+1}$, we have (C.1) holds for all such unit vectors $z$, therefore also for all vectors $z \in \mathbb{R}^{k+1}$.[18] In particular, choosing $z = (\mathsf{out}_1(x), 1)$ finishes the proof.

(e) We write
$$\mathbf{A}D_{\mathbf{V},\mathbf{W}}\mathbf{V}(\mathsf{out}_1(x), 1) - \mathbf{A}D_{\mathbf{V},\mathbf{W}}(\mathbf{V}^{(0)} + \mathbf{V})(\mathsf{out}_1(x), 1)$$
$$= -\mathbf{A}D_{\mathbf{V}^{(0)},\mathbf{W}}\mathbf{V}^{(0)}(\mathsf{out}_1(x), 1) + \mathbf{A}(D_{\mathbf{V}^{(0)},\mathbf{W}} - D_{\mathbf{V},\mathbf{W}})\mathbf{V}^{(0)}(\mathsf{out}_1(x), 1)$$

---

[18] More formally, this requires one to construct a set $\{z_1, z_2, \ldots\} \subset \mathbb{R}^{k+1}$ of $\varepsilon^{-\Omega(k)}$ unit vectors so that each unit vector is at most $\varepsilon$-close to some point in this set with $\varepsilon = 1/\mathsf{poly}(m)$. Then, one can derive that as long as $\varepsilon$ is sufficiently small, for each $i$, with probability at least $1 - e^{-\Omega(m^{2/3})}$ inequality (C.1) holds for all unit vectors $z$ with $\|z - z_i\|_2 \leq \varepsilon$. Taking union bond over all $z_i$ in this set finishes the argument.

Let us denote by $z = (\mathsf{out}_1(x), 1)$. Again, suppose for now that $z \in \mathbb{R}^{k+1}$ is a fixed vector that does not depend on $\mathbf{W}^{(0)}$ or $\mathbf{W}$.

Then, for the first term, we have $\mathbf{A}D_{\mathbf{V}^{(0)},\mathbf{W}}\mathbf{V}^{(0)}z = \mathbf{A}\sigma(\mathbf{V}^{(0)}z)$ and by by concentration of chi-square distribution we have $\|\mathbf{V}^{(0)}z\|_2 \leq O(\sigma_v\|z\|_2)$ with probability at least $1 - e^{-\Omega(m)}$, and then using the randomness of $\mathbf{A}$ and applying chi-square concentration again (see Proposition B.1), we have with probability at least $1 - e^{-\Omega(k \log^2 m)}$,
$$\left\|\mathbf{A}D_{\mathbf{V}^{(0)},\mathbf{W}}\mathbf{V}^{(0)}z\right\|_2 \leq \widetilde{O}(\sqrt{k}/\sqrt{m}) \cdot O(\sigma_v\|z\|_2) \enspace.$$

For the second term, invoking Proposition B.4, we have
$$\left\|(D_{\mathbf{V},\mathbf{W}} - D_{\mathbf{V}^{(0)},\mathbf{W}})\mathbf{V}^{(0)}z\right\|_2 \leq \|\mathbf{V}z\|_2 \leq \tau_v \cdot \|z\|_2$$

Recall for every $s$-sparse vectors $y$, it satisfies $\|\mathbf{A}y\|_2 \leq \widetilde{O}(\frac{\sqrt{s}}{\sqrt{m}}) \cdot \|y\|_2$ with probability at least $1 - e^{-\widetilde{\Omega}(s)}$ (see Proposition B.2). This implies
$$\left\|\mathbf{A}(D_{\mathbf{V},\mathbf{W}} - D_{\mathbf{V}^{(0)},\mathbf{W}})\mathbf{V}^{(0)}z\right\|_2 \leq \widetilde{O}(\frac{\sqrt{s}}{\sqrt{m}}\tau_v\|z\|_2)$$

for $s = O\left(\left(\frac{\tau_v}{\sigma_v}\right)^{2/3} \cdot m\right)$. Combining the two bounds above, we have for every fixed $z \in \mathbb{R}^{k+1}$,
$$\left\|\mathbf{A}D_{\mathbf{V},\mathbf{W}}\mathbf{V}^{(0)}z\right\|_2 = \left\|\mathbf{A}D_{\mathbf{V},\mathbf{W}}\mathbf{V}z - \mathbf{A}D_{\mathbf{V},\mathbf{W}}(\mathbf{V}^{(0)} + \mathbf{V})z\right\|_2$$
$$\leq \widetilde{O}\left((\tau_v(\tau_v/\sigma_v)^{1/3} + \sqrt{k}\sigma_v/\sqrt{m})\|z\|_2\right) \leq \widetilde{O}\left(\|z\|_2\tau_v(\tau_v/\sigma_v)^{1/3}\right) \enspace.$$

with probability at least $1 - e^{-\Omega(k \log^2 m)}$. Finally, because this confidence is sufficiently small, one can take an $\varepsilon$-net over all possible vectors $z \in \mathbb{R}^{k+1}$ and derive the above bound for all vectors $z$. In particular, choosing $z = (\mathsf{out}_1(x), 1)$ finishes the proof.

(f) This is a byproduct of the proof of Lemma C.1e.

(g) With high probability
$$\|\mathbf{A}D_{\mathbf{V},\mathbf{W}}\mathbf{V}(\mathsf{out}_1(x), 1)\|_2 \leq \|\mathbf{A}\|_2 \cdot \|\mathbf{V}\|_2 \cdot (1 + \|\mathsf{out}_1(x)\|_2) \leq O(\tau_v) \cdot (1 + \|\mathsf{out}_1(x)\|_2)$$

Combining this with Lemma C.1e gives the proof.

$\square$

## C.2 Existantial

In this subsection, we prove the existence of matrices $\mathbf{W}^*, \mathbf{V}^*$ with $\|\mathbf{W}^*\|_F \leq \frac{\tau_w}{10}$ and $\|\mathbf{V}^*\|_F \leq \frac{\tau_v}{10}$ satisfying $\mathbf{A}D_{\mathbf{W}^{(0)}}\mathbf{W}^*(x, 1) \approx \mathcal{F}(x)$ and $\mathbf{A}D_{\mathbf{V}^{(0)},\mathbf{W}}\mathbf{V}^*(\mathsf{out}_1(x), 1) \approx \alpha \mathcal{G}(\mathsf{out}_1(x))$.

This existential proof relies on an "indicator to function" lemma that was used in prior work [4]; however, for the purpose of this paper we have to revise it to include a trainable bias term (or equivalently, to support vectors of the form $(x, 1)$). We treat that carefully in Appendix E.

**Lemma C.2.** *Suppose $\alpha \in (0, 1)$ and $\widetilde{\alpha} = \frac{\alpha}{k(p_\mathcal{F}\mathfrak{C}_\mathfrak{s}(\mathcal{F}) + p_\mathcal{G}\mathfrak{C}_\mathfrak{s}(\mathcal{G}))}$, there exist $M = \mathsf{poly}(\mathfrak{C}_{\widetilde{\alpha}}(\mathcal{F}), \mathfrak{C}_{\widetilde{\alpha}}(\mathcal{G}), \widetilde{\alpha}^{-1})$ satisfying that for every $m \geq M$, with high probability over $\mathbf{A}, \mathbf{W}^{(0)}, \mathbf{V}^{(0)}$, one can construct $\mathbf{W}^* \in \mathbb{R}^{m \times (d+1)}$ and $\mathbf{V}^* \in \mathbb{R}^{m \times (k+1)}$ with*
$$\|\mathbf{W}^*\|_F \leq \frac{\tau_w}{10} \stackrel{\mathrm{def}}{=} \widetilde{O}(kp_\mathcal{F}\mathfrak{C}_\mathfrak{s}(\mathcal{F})) \text{ and } \|\mathbf{V}^*\|_F \leq \frac{\tau_v}{10} \stackrel{\mathrm{def}}{=} \widetilde{O}(\widetilde{\alpha}kp_\mathcal{G}\mathfrak{C}_\mathfrak{s}(\mathcal{G}))$$

*satisfying*

(a) $\mathbb{E}_{(x,y)\sim\mathcal{D}}\left[\|\mathbf{A}D_{\mathbf{W}^{(0)}}\mathbf{W}^*(x,1) - \mathcal{F}(x)\|_2^2\right] \leq \widetilde{\alpha}^4$;

(b) for all $x$ and $\mathbf{W}$, $\left\|\mathbf{A}D_{\mathbf{V}^{(0)},\mathbf{W}}\mathbf{V}^*(\mathsf{out}_1(x),1) - \alpha\mathcal{G}(\mathsf{out}_1(x))\right\|_2 \leq \widetilde{\alpha}^2 \cdot \|(\mathsf{out}_1(x),1)\|_2$.

*Proof.*

(a) For each $r \in [k]$, we have $[\mathbf{A}D_{\mathbf{W}^{(0)}}\mathbf{W}^*(x,1)]_k = \sum_{i\in[m]} a_{r,i}\mathbb{1}_{\langle w_i^{(0)},(x,1)\rangle\geq 0}\langle w_i^*,(x,1)\rangle$. By applying Lemma E.1 (which is a simple modification on top of the existential result from [4]), we can construct matrix $\mathbf{W}^*$ satisfying $\|\mathbf{A}D_{\mathbf{W}^{(0)}}\mathbf{W}^*(x,1) - \mathcal{F}(x)\|_2 \leq \widetilde{\alpha}^2 \cdot \|(x,1)\|_2$ for each $x \in \mathbb{R}^d$ with probability at least $1 - e^{-\Omega(\sqrt{m})}$. This translates to an expected guarantee with respect to $(x,y)\sim\mathcal{D}$.

(b) For each $r \in [k]$, we have

$$[\mathbf{A}D_{\mathbf{V}^{(0)},\mathbf{W}}\mathbf{V}^*(\mathsf{out}_1(x),1)]_k = \sum_{i\in[m]} a_{r,i}\mathbb{1}_{\langle v_i^{(0)},(\mathsf{out}_1(x),1)\rangle\geq 0}\langle v_i^*,(\mathsf{out}_1(x),1)\rangle.$$

Now, applying Lemma E.1 again, we can construct matrix $\mathbf{V}^*$ satisfying for each $z \in \mathbb{R}^k$ with probability at least $1 - e^{-\Omega(\sqrt{m})}$:

$$\sum_{r\in[k]}\left|\sum_{i\in[m]} a_{r,i}\mathbb{1}_{\langle v_i^{(0)},(z,1)\rangle\geq 0}\langle v_i^*,(z,1)\rangle - \alpha\mathcal{G}(z)\right| \leq \frac{\widetilde{\alpha}^2}{2} \cdot \|(z,1)\|_2.$$

By applying a careful $\varepsilon$-net argument and using $m \geq \mathsf{poly}(k)$,[19] this translates to, with probability at least $1 - e^{-\Omega(\sqrt{m})}$, for *all* vectors $z \in \mathbb{R}^k$.

$$\sum_{r\in[k]}\left|\sum_{i\in[m]} a_{r,i}\mathbb{1}_{\langle v_i^{(0)},(z,1)\rangle\geq 0}\langle v_i^*,(z,1)\rangle - \alpha\mathcal{G}(z)\right| \leq \widetilde{\alpha}^2 \cdot \|(z,1)\|_2 \quad (C.2)$$

Finally, choosing $z = \mathsf{out}_1(x)$ finishes the proof. $\square$

Next, we can combine coupling and existential lemmas:

**Lemma C.3.** *Under the assumptions of Lemma C.1 and Lemma C.2, we have*

(a) $\mathbb{E}_{(x,y)\sim\mathcal{D}}\left[\|\mathbf{A}D_{\mathbf{W}}\mathbf{W}^*(x,1) - \mathcal{F}(x)\|_2^2\right] \leq O(\widetilde{\alpha}^4)$

(b) $\forall x \in \mathbb{R}^d$, $\|\mathbf{A}D_{\mathbf{V},\mathbf{W}}\mathbf{V}^*(\mathsf{out}_1(x),1) - \widetilde{\alpha}\mathcal{G}(\mathsf{out}_1(x))\|_2 \leq \left(\widetilde{\alpha}^2 + O(\tau_v(\tau_v/\sigma_v)^{1/3})\right) \cdot \|(\mathsf{out}_1(x),1)\|_2$

(c) $\mathbb{E}_{(x,y)\sim\mathcal{D}}\left[\|\mathbf{A}D_{\mathbf{W}}(\mathbf{W}^* - \mathbf{W})(x,1) - (\mathcal{F}(x) - \mathsf{out}_1(x))\|_2^2\right] \leq O(\widetilde{\alpha}^4)$

*Proof.*

(a) For every $s$-sparse vectors $y$, it satisfies $\|\mathbf{A}y\|_2 \leq \widetilde{O}(\frac{\sqrt{s}}{\sqrt{m}}) \cdot \|y\|_2$ with high probability (see Proposition B.2). We also have $\|\mathbf{W}^*(x,1)\|_2 \leq O(\|\mathbf{W}^*\|_F) \leq O(\tau_w)$. Therefore, $\|\mathbf{A}(D_{\mathbf{W}^{(0)}} - D_{\mathbf{W}})\mathbf{W}^*(x,1)\| \leq O(\sqrt{s}\tau_w/\sqrt{m})$ where $s$ is the maximum sparsity of $D_{\mathbf{W}^{(0)}} - D_{\mathbf{W}}$, which satisfies $s = O((\tau_w/\sigma_w)^{2/3}m^{2/3})$ by Lemma C.1a. This, combining with Lemma C.2a gives

$$\mathbb{E}_{(x,y)\sim\mathcal{D}}\left[\|\mathbf{A}D_{\mathbf{W}}\mathbf{W}^*(x,1) - \mathcal{F}(x)\|_2^2\right] \leq 2\widetilde{\alpha}^2 + O(\tau_w(\tau_w/\sigma_w)^{1/3}/m^{1/6})^2 \leq O(\widetilde{\alpha}^4).$$

---

[19] This is a bit non-trivial to derive, because one has to argue that if $z$ changes a little bit (i.e., by $1/\mathsf{poly}(m)$), then according to Lemma C.1d, the number of sign changes in $\{\mathbb{1}_{\langle v_i^{(0)},(z,1)\rangle\geq 0}\}_{i\in[m]}$ is $o(m)$, and thus the interested quantity changes by at most $1/\mathsf{poly}(m)$.

(b) Again, for every $s$-sparse vectors $y$, it satisfies $\|\mathbf{A}y\|_2 \leq \widetilde{O}(\frac{\sqrt{s}}{\sqrt{m}}) \cdot \|y\|_2$ with high probability. We also have $\|\mathbf{V}^*(\mathsf{out}_1(x), 1)\|_2 \leq O(\|\mathbf{V}^*\|_F) \cdot \|(\mathsf{out}_1(x), 1)\|_2 \leq O(\tau_v) \cdot \|(\mathsf{out}_1(x), 1)\|_2$. Therefore,
$$\|\mathbf{A}(D_{\mathbf{V}^{(0)},\mathbf{W}} - D_{\mathbf{V},\mathbf{W}})\mathbf{V}^*(\mathsf{out}_1(x), 1)\| \leq O(\sqrt{s}\tau_v/\sqrt{m}) \cdot \|(\mathsf{out}_1(x), 1)\|_2$$
where $s$ is the maximum sparsity of $D_{\mathbf{V}^{(0)},\mathbf{W}} - D_{\mathbf{V},\mathbf{W}}$, which satisfies $s = O((\tau_v/\sigma_v)^{2/3}m)$ by Lemma C.1d. This, combining with Lemma C.2b gives
$$\mathop{\mathbb{E}}_{(x,y)\sim\mathcal{D}}\left[\|\mathbf{A}D_{\mathbf{W}}\mathbf{W}^*(x, 1) - \mathcal{F}(x)\|_2\right] \leq \left(\widetilde{\alpha}^2 + O(\tau_v(\tau_v/\sigma_v)^{1/3})\right) \cdot \|(\mathsf{out}_1(x), 1)\|_2 \ .$$

(c) This combines Lemma C.1b and Lemma C.3a, together with our sufficiently large choice of $m$. □

## C.3 Optimization

In this subsection we give some structural results that shall be later used in the optimization step. The first fact gives an explicit formula of the gradient.

**Fact C.4.** *When* $\mathsf{Obj}(\mathbf{W}, \mathbf{V}; (x, y)) = \frac{1}{2}\|y - \mathsf{out}(\mathbf{W}, \mathbf{V}; x)\|_2^2$, *we can write its gradient as follows.*
$$\langle \nabla_{\mathbf{W},\mathbf{V}}\mathsf{Obj}(\mathbf{W}, \mathbf{V}; (x, y)), (-\mathbf{W}', -\mathbf{V}')\rangle = \langle y - \mathsf{out}(x), f(\mathbf{W}'; x) + g(\mathbf{V}'; x)\rangle$$
*where*
$$f(\mathbf{W}'; x) = \mathbf{A}D_{\mathbf{V},\mathbf{W}}(\mathbf{V}^{(0)} + \mathbf{V})\left(\mathbf{A}D_{\mathbf{W}}\mathbf{W}'(x, 1), 0\right) + \mathbf{A}D_{\mathbf{W}}\mathbf{W}'(x, 1)$$
$$g(\mathbf{V}'; x) = \mathbf{A}D_{\mathbf{V},\mathbf{W}}\mathbf{V}'(\mathsf{out}_1(x), 1)$$

The next claim gives simple upper bound on the norm of the gradient.

**Claim C.5.** *For all* $(x, y)$ *in the support of* $\mathcal{D}$, *with high probability over* $\mathbf{A}, \mathbf{W}^{(0)}, \mathbf{V}^{(0)}$, *we have that for all* $\mathbf{W}, \mathbf{V}$ *satisfying* $\|\mathbf{W}\|_F \leq \tau_w$ *and* $\|\mathbf{V}\|_F \leq \tau_v$, *it holds that*
$$\|\nabla_{\mathbf{W}}\mathsf{Obj}(\mathbf{W}, \mathbf{V}; (x, y))\|_F \leq \|y - \mathsf{out}(x)\|_2 \cdot O(\sigma_v + 1)$$
$$\|\nabla_{\mathbf{V}}\mathsf{Obj}(\mathbf{W}, \mathbf{V}; (x, y))\|_F \leq \|y - \mathsf{out}(x)\|_2 \cdot O(\tau_w + 1) \ .$$

*Proof.* For the gradient in $\mathbf{W}$, we derive using the gradient formula Fact C.4 that
$$\|\nabla_{\mathbf{W}}\mathsf{Obj}(\mathbf{W}, \mathbf{V}; (x, y))\|_F = \left\|(x, 1)(y - \mathsf{out}(x))^\top \left(\mathbf{A}D_{\mathbf{V},\mathbf{W}}(\mathbf{V}^{(0)} + \mathbf{V})(\mathbf{A}D_{\mathbf{W}}, 0) + \mathbf{A}D_{\mathbf{W}}\right)\right\|_F$$
$$= \|(x, 1)\|_2 \cdot \left\|(y - \mathsf{out}(x))^\top \left(\mathbf{A}D_{\mathbf{V},\mathbf{W}}(\mathbf{V}^{(0)} + \mathbf{V})(\mathbf{A}D_{\mathbf{W}}, 0) + \mathbf{A}D_{\mathbf{W}}\right)\right\|_2$$
$$\leq 2\|y - \mathsf{out}(x)\|_2 \cdot \left\|\mathbf{A}D_{\mathbf{V},\mathbf{W}}(\mathbf{V}^{(0)} + \mathbf{V})(\mathbf{A}D_{\mathbf{W}}, 0) + \mathbf{A}D_{\mathbf{W}}\right\|_2$$
$$\leq \|y - \mathsf{out}(x)\|_2 \cdot O(\sigma_v + 1) \ .$$

Above, the last inequality uses $\|\mathbf{A}\|_2 \leq O(1)$ and $\|\mathbf{V}^{(0)}\|_2 \leq O(\sigma_v)$ with high probability (using random matrix theory, see Proposition B.2), as well as $\tau_v \leq \sigma_v$. Similarly, using the gradient formula Fact C.4, we derive that
$$\|\nabla_{\mathbf{V}}\mathsf{Obj}(\mathbf{W}, \mathbf{V}; (x, y))\|_F = \left\|(\mathsf{out}_1(x), 1)(y - \mathsf{out}(x))^\top \mathbf{A}D_{\mathbf{V},\mathbf{W}}\right\|_F$$
$$= \|(\mathsf{out}_1(x), 1)\|_2 \cdot \left\|(y - \mathsf{out}(x))^\top \mathbf{A}D_{\mathbf{V},\mathbf{W}}\right\|_2$$
$$\leq \|y - \mathsf{out}(x)\|_2 \cdot O(\tau_w + 1) \cdot O(1)$$
where the last inequality uses Lemma C.1c and $\|\mathbf{A}\|_2 \leq O(1)$. □

The next claim gives a careful approximation to $f(\mathbf{W}^* - \mathbf{W}; x) + g(\mathbf{V}^* - \mathbf{V}; x)$, which according to Fact C.4 is related to the correlation between the gradient direction and $(\mathbf{W} - \mathbf{W}^*, \mathbf{V} - \mathbf{V}^*)$.

**Claim C.6.** *In the same setting as Lemma C.1 and Lemma C.2, suppose we set parameters according to Table 1. Then, we can write*

$$f(\mathbf{W}^* - \mathbf{W}; x) + g(\mathbf{V}^* - \mathbf{V}; x) = \mathcal{H}(x) - \mathsf{out}(x) + Err$$

*with*
$$\mathop{\mathbb{E}}_{(x,y)\sim\mathcal{D}} \|Err\|_2^2 \leq O(\tau_v + \alpha \mathfrak{L}_{\mathcal{G}})^2 \cdot \mathop{\mathbb{E}}_{(x,y)\sim\mathcal{D}} \|\mathcal{H}(x) - \mathsf{out}(x)\|_2^2$$
$$+ O\left(\widetilde{\alpha}^2 + \tau_v^2(1 + \mathfrak{B}_{\mathcal{F}}) + \alpha\tau_v \mathfrak{L}_{\mathcal{G}}(\mathfrak{B}_{\mathcal{F}} + 1)\right)^2 .$$

*and for every $(x,y) \sim \mathcal{D}$, with high probability $\|Err\|_2 \leq O(\tau_w)$.*

*Proof of Claim C.6.*

$$f(\mathbf{W}^* - \mathbf{W}; x) + g(\mathbf{V}^* - \mathbf{V}; x)$$
$$= \mathbf{A}D_{\mathbf{V},\mathbf{W}}(\mathbf{V}^{(0)} + \mathbf{V})\left(\mathbf{A}D_{\mathbf{W}}(\mathbf{W}^* - \mathbf{W})(x,1),0\right) + \mathbf{A}D_{\mathbf{W}}(\mathbf{W}^* - \mathbf{W})(x,1) + \mathbf{A}D_{\mathbf{V},\mathbf{W}}(\mathbf{V}^* - \mathbf{V})(\mathsf{out}_1(x),1)$$
$$= \underbrace{\mathbf{A}D_{\mathbf{V},\mathbf{W}}(\mathbf{V}^{(0)} + \mathbf{V})\left(\mathbf{A}D_{\mathbf{W}}(\mathbf{W}^* - \mathbf{W})(x,1),0\right)}_{\clubsuit} + \underbrace{\left(\mathbf{A}D_{\mathbf{W}}\mathbf{W}^*(x,1) + \mathbf{A}D_{\mathbf{V},\mathbf{W}}\mathbf{V}^*(\mathsf{out}_1(x),1)\right)}_{\spadesuit}$$
$$- \underbrace{\left(\mathbf{A}D_{\mathbf{W}}\mathbf{W}(x,1) + \mathbf{A}D_{\mathbf{V},\mathbf{W}}\mathbf{V}(\mathsf{out}_1(x),1)\right)}_{\diamondsuit}$$

We treat the three terms separately.

- For the $\clubsuit$ term, under expectation over $(x,y) \sim \mathcal{D}$,
$$\|\clubsuit\|_2^2 \leq \left(\|\mathbf{A}D_{\mathbf{V},\mathbf{W}}\mathbf{V}^{(0)}\|_2 + \|\mathbf{A}\|_2^2\|\mathbf{V}\|_2^2\right) \|\mathbf{A}D_{\mathbf{W}}(\mathbf{W}^* - \mathbf{W})x\|_2^2$$
$$\leq O(1) \cdot O(\tau_v)^2 \cdot \left(\|\mathcal{F}(x) - \mathsf{out}_1(x)\|_2^2 + O(\widetilde{\alpha}^2)\right)$$

where the last inequality uses Lemma C.1f and Lemma C.3c, together with $\tau_v \leq \frac{1}{\mathsf{polylog}(m)} \sigma_v$.

- For the $\spadesuit$ term, under expectation over $(x,y) \sim \mathcal{D}$,
$$\|\spadesuit - (\mathcal{F}(x) + \alpha\mathcal{G}(\mathcal{F}(x)))\|_2^2 \leq O(\widetilde{\alpha}^2 + \tau_v(\tau_v/\sigma_v)^{1/3})^2 \cdot (\|\mathsf{out}_1(x)\|_2 + 1)^2$$
$$+ O(\alpha\mathfrak{L}_{\mathcal{G}})^2 \|\mathcal{F}(x) - \mathsf{out}_1(x)\|_2^2$$
$$\leq O(\tau_v^2)^2 \cdot (\|\mathsf{out}_1(x)\|_2 + 1)^2 + O(\alpha\mathfrak{L}_{\mathcal{G}})^2 \|\mathcal{F}(x) - \mathsf{out}_1(x)\|_2^2$$

where the first inequality uses Lemma C.3a and Lemma C.3b, as well as the Lipscthiz continuity of $\mathcal{G}(x)$ (which satisfies $\|\mathcal{G}(x) - \mathcal{G}(y)\| \leq \mathfrak{L}_{\mathcal{G}}\|x-y\|$); and the second inequality uses $\frac{1}{\sigma_v} \leq \tau_v^2$ and the definition of $\widetilde{\alpha}$.

- For the $\diamondsuit$ term, under expectation over $(x,y) \sim \mathcal{D}$,
$$\|\diamondsuit - \mathsf{out}(x)\|_2^2 \leq O\left((\|\mathsf{out}_1(x)\|_2 + 1)\tau_v^2\right)^2$$

where the inequality uses Lemma C.1b, Lemma C.1e and $\frac{1}{\sigma_v} \leq \tau_v^2$.

In sum, we have
$$Err \stackrel{\text{def}}{=} f(\mathbf{W}^* - \mathbf{W}; x) + g(\mathbf{V}^* - \mathbf{V}; x) - (\mathcal{F}(x) + \alpha\mathcal{G}(\mathcal{F}(x))) - \mathsf{out}(x)$$

satisfies
$$\mathop{\mathbb{E}}_{(x,y)\sim\mathcal{D}} \|Err\|_2^2 \leq \mathop{\mathbb{E}}_{(x,y)\sim\mathcal{D}} \left[O(\tau_v + \alpha\mathfrak{L}_{\mathcal{G}})^2 \cdot \|\mathcal{F}(x) - \mathsf{out}_1(x)\|_2^2 + O\left(\widetilde{\alpha}^2 + (\|\mathsf{out}_1(x)\|_2 + 1)\tau_v^2\right)^2\right] .$$

Combining this with Claim C.7, and using $\|\mathsf{out}_1(x)\|_2 \leq \|\mathsf{out}_1(x) - \mathcal{F}(x)\|_2 + \mathfrak{B}_\mathcal{F}$, we have

$$\mathop{\mathbb{E}}_{(x,y)\sim\mathcal{D}} \|Err\|_2^2 \leq O(\tau_v + \alpha\mathfrak{L}_\mathcal{G})^2 \cdot \mathop{\mathbb{E}}_{(x,y)\sim\mathcal{D}} \|\mathcal{H}(x) - \mathsf{out}(x)\|_2^2 + O\left(\widetilde{\alpha}^2 + (1+\mathfrak{B}_\mathcal{F})\tau_v^2\right)^2$$
$$+ O(\tau_v + \alpha\mathfrak{L}_\mathcal{G})^2 \cdot (\tau_v(\mathfrak{B}_\mathcal{F} + 1) + \alpha\mathfrak{B}_{\mathcal{F}\circ\mathcal{G}})^2 .$$

Using $\mathfrak{B}_{\mathcal{F}\circ\mathcal{G}} \leq \sqrt{k}p_\mathcal{F}\mathfrak{C}_\mathfrak{s}(\mathcal{F})\mathfrak{B}_\mathcal{F} \leq \frac{\tau_v}{\alpha}\mathfrak{B}_\mathcal{F}$ (see Fact A.3), we finish the bound on $\mathbb{E}_{(x,y)\sim\mathcal{D}} \|Err\|_2^2$.

As for the absolute value bound, one can naively derive that with high probability $\|f(\mathbf{W}^* - \mathbf{W}; x)\|_2 \leq O(\tau_w)$, $\|g(\mathbf{V}^* - \mathbf{V}; x)\|_2 \leq O(\tau_w\tau_v)$, $\|\mathcal{H}(x)\|_2 \leq \mathfrak{B}_\mathcal{F} + \alpha\mathfrak{B}_{\mathcal{F}\circ\mathcal{G}}$, and $\|\mathsf{out}(\tau_w)\|_2 \leq O(\tau_w)$ (by Lemma C.1c and C.1g). Combining them with $\mathfrak{B}_\mathcal{F} \leq \sqrt{k}p_\mathcal{F}\mathfrak{C}_\mathfrak{s}(\mathcal{F}) \leq \tau_w$ and $\alpha\mathfrak{B}_{\mathcal{F}\circ\mathcal{G}} \leq \alpha(\mathfrak{B}_\mathcal{F}\mathfrak{L}_\mathcal{G} + \mathfrak{C}_\mathfrak{s}(\mathcal{G})) \leq \frac{1}{kp_\mathcal{G}\mathfrak{C}_\mathfrak{s}(\mathcal{G})}(\mathfrak{B}_\mathcal{F}\mathfrak{L}_\mathcal{G} + \mathfrak{C}_\mathfrak{s}(\mathcal{G})) \leq \tau_w$ finishes the proof. □

Finally, we state a simple claim that bounds the norm of $\|\mathsf{out}_1(x) - \mathcal{F}(x)\|_2$ given the norm of $\|\mathsf{out}(x) - \mathcal{H}(x)\|_2$.

**Claim C.7.** *In the same setting as Lemma C.1, if we additionally have $\tau_v \leq \frac{1}{\mathsf{polylog}(m)}$, for every fixed $x$, with high probability over $\mathbf{A}, \mathbf{W}^{(0)}, \mathbf{V}^{(0)}$,*

$$\|\mathsf{out}_1(x) - \mathcal{F}(x)\|_2 \leq 2\|\mathsf{out}(x) - \mathcal{H}(x)\|_2 + \widetilde{O}(\tau_v(\mathfrak{B}_\mathcal{F}+1) + \alpha\mathfrak{B}_{\mathcal{F}\circ\mathcal{G}}) .$$

*Proof.* We can rewrite

$$\mathsf{out}_1(x) - \mathcal{F}(x) = (\mathsf{out}(x) - \mathcal{H}(x)) - \left(\mathbf{A}D_{\mathbf{V},\mathbf{W}}(\mathbf{V}^{(0)} + \mathbf{V})(\mathsf{out}_1(x), 1)\right) + \alpha\mathcal{G}(\mathcal{F}(x)) .$$

Using Lemma C.1g we have $\|\mathbf{A}D_{\mathbf{V},\mathbf{W}}(\mathbf{V}^{(0)} + \mathbf{V})(\mathsf{out}_1(x), 1)\| \leq \widetilde{O}(\tau_v(\|\mathsf{out}_1(x)\|_2 + 1))$, and using the boundedness we have $\|\mathcal{G}(\mathcal{F}(x))\|_2 \leq \mathfrak{B}_{\mathcal{F}\circ\mathcal{G}}$. We also have $\|\mathsf{out}_1(x)\|_2 \leq \|\mathsf{out}_1(x) - \mathcal{F}(x)\|_2 + \mathfrak{B}_\mathcal{F}$. Together, we have

$$\|\mathsf{out}_1(x) - \mathcal{F}(x)\|_2 \leq \|\mathsf{out}(x) - \mathcal{H}(x)\|_2 + \widetilde{O}(\tau_v(\|\mathsf{out}_1(x) - \mathcal{F}(x)\|_2 + \mathfrak{B}_\mathcal{F} + 1)) + \alpha\mathfrak{B}_{\mathcal{F}\circ\mathcal{G}} .$$

Using $\tau_v \leq \frac{1}{\mathsf{polylog}(m)}$ we finish the proof. □

### C.4 Proof of Theorem 1

**Theorem 1.** *Under Concept 1 or Concept 2, for every $\alpha \in \left(0, \widetilde{\Theta}(\frac{1}{kp_\mathcal{G}\mathfrak{C}_\mathfrak{s}(\mathcal{G})})\right)$ and $\delta \geq \mathsf{OPT} + \widetilde{\Theta}\left(\alpha^4(kp_\mathcal{G}\mathfrak{C}_\mathfrak{s}(\mathcal{G}))^4(1+\mathfrak{B}_\mathcal{F})^2\right)$. There exist $M = \mathsf{poly}(\mathfrak{C}_\alpha(\mathcal{F}), \mathfrak{C}_\alpha(\mathcal{G}), p_\mathcal{F}, \alpha^{-1})$ satisfying that for every $m \geq M$, with high probability over $\mathbf{A}, \mathbf{W}^{(0)}, \mathbf{V}^{(0)}$, for a wide range of random initialization parameters $\sigma_w, \sigma_v$ (see Table 1), choosing*

$$T = \widetilde{\Theta}\left(\frac{(kp_\mathcal{F}\mathfrak{C}_\mathfrak{s}(\mathcal{F}))^2}{\min\{1,\delta^2\}}\right) \quad \eta_w = \widetilde{\Theta}(\min\{1,\delta\}) \quad \eta_v = \eta_w \cdot \widetilde{\Theta}\left(\frac{\alpha p_\mathcal{G}\mathfrak{C}_\mathfrak{s}(\mathcal{G})}{p_\mathcal{F}\mathfrak{C}_\mathfrak{s}(\mathcal{F})}\right)^2$$

*With high probability, the SGD algorithm satisfies*

$$\frac{1}{T}\sum_{t=0}^{T-1} \mathop{\mathbb{E}}_{(x,y)\sim\mathcal{D}} \|\mathcal{H}(x) - \mathsf{out}(\mathbf{W}_t, \mathbf{V}_t; x)\|_2^2 \leq O(\delta) .$$

*Proof of Theorem 1.* We first assume that throughout the SGD algorithm, it satisfies

$$\|\mathbf{W}_t\|_F \leq \tau_w \quad \text{and} \quad \|\mathbf{V}_t\|_F \leq \tau_v . \tag{C.3}$$

We shall prove in the end that (C.3) holds throughout the SGD algorithm.

On one hand, using Claim C.6, at any point $\mathbf{W}_t, \mathbf{V}_t$, we have

$$\langle \nabla_{\mathbf{W},\mathbf{V}} \mathsf{Obj}(\mathbf{W}_t, \mathbf{V}_t; (x_t, y_t)), (\mathbf{W}_t - \mathbf{W}^*, \mathbf{V}_t - \mathbf{V}^*)) \rangle$$
$$= \langle y_t - \mathsf{out}(\mathbf{W}_t, \mathbf{V}_t; x_t), \mathcal{H}(x_t) - \mathsf{out}(\mathbf{W}_t, \mathbf{V}_t; x_t) + Err_t \rangle$$
$$\geq \frac{1}{2}\|\mathcal{H}(x_t) - \mathsf{out}(\mathbf{W}_t, \mathbf{V}_t; x_t)\|_2^2 - 2\|Err_t\|_2^2 - 2\|\mathcal{H}(x_t) - y_t\|_2^2$$

where $Err_t$ comes from Claim C.6. On the other hand, using $\mathbf{W}_{t+1} = \mathbf{W}_t - \eta_w \nabla_{\mathbf{W}} \mathsf{Obj}(\mathbf{W}_t, \mathbf{V}_t; (x_t, y_t))$ and $\mathbf{V}_{t+1} = \mathbf{V}_t - \eta_v \nabla_{\mathbf{V}} \mathsf{Obj}(\mathbf{W}_t, \mathbf{V}_t; (x_t, y_t))$, we have

$$\langle \nabla_{\mathbf{W},\mathbf{V}} \mathsf{Obj}(\mathbf{W}_t, \mathbf{V}_t; (x_t, y_t)), (\mathbf{W} - \mathbf{W}^*, \mathbf{V} - \mathbf{V}^*)) \rangle$$
$$= \underbrace{\frac{\eta_w}{2}\|\nabla_{\mathbf{W}} \mathsf{Obj}(\mathbf{W}_t, \mathbf{V}_t; (x_t, y_t))\|_F^2 + \frac{\eta_v}{2}\|\nabla_{\mathbf{V}} \mathsf{Obj}(\mathbf{W}_t, \mathbf{V}_t; (x_t, y_t))\|_F^2}_{\heartsuit}$$
$$+ \frac{1}{2\eta_w}\|\mathbf{W}_t - \mathbf{W}^*\|_F^2 - \frac{1}{2\eta_w}\|\mathbf{W}_{t+1} - \mathbf{W}^*\|_F^2 + \frac{1}{2\eta_v}\|\mathbf{V}_t - \mathbf{V}^*\|_F^2 - \frac{1}{2\eta_v}\|\mathbf{V}_{t+1} - \mathbf{V}^*\|_F^2$$

Recall from Claim C.5,

$$\heartsuit \leq O(\eta_w + \eta_v \tau_w^2) \cdot \|y_t - \mathsf{out}(\mathbf{W}_t, \mathbf{V}_t; x_t)\|_2^2 \leq O(\eta_w + \eta_v \tau_w^2) \cdot \left(\|\mathcal{H}(x_t) - \mathsf{out}(\mathbf{W}_t, \mathbf{V}_t; x_t)\|_2^2 + \|\mathcal{H}(x_t) - y_t\|_2^2\right)$$

Therefore, as long as $O(\eta_w + \eta_v \tau_w^2) \leq 0.1$, it satisfies

$$\frac{1}{4}\|\mathcal{H}(x_t) - \mathsf{out}(\mathbf{W}_t, \mathbf{V}_t; x_t)\|_2^2 \leq 2\|Err_t\|_2^2 + 4\|\mathcal{H}(x_t) - y_t\|_2^2 + \frac{1}{2\eta_w}\|\mathbf{W}_t - \mathbf{W}^*\|_F^2 - \frac{1}{2\eta_w}\|\mathbf{W}_{t+1} - \mathbf{W}^*\|_F^2$$
$$+ \frac{1}{2\eta_v}\|\mathbf{V}_t - \mathbf{V}^*\|_F^2 - \frac{1}{2\eta_v}\|\mathbf{V}_{t+1} - \mathbf{V}^*\|_F^2$$

After telescoping for $t = 0, 1, \ldots, T_0 - 1$,

$$\frac{\|\mathbf{W}_{T_0} - \mathbf{W}^*\|_F^2}{2\eta_w T_0} + \frac{\|\mathbf{W}_{T_0} - \mathbf{V}^*\|_F^2}{2\eta_v T_0} + \frac{1}{2T_0} \sum_{t=0}^{T_0-1} \|\mathcal{H}(x_t) - \mathsf{out}(\mathbf{W}_t, \mathbf{V}_t; x_t)\|_2^2$$
$$\leq \frac{\|\mathbf{W}^*\|_F^2}{2\eta_w T_0} + \frac{\|\mathbf{V}^*\|_F^2}{2\eta_v T_0} + \frac{O(1)}{T_0} \sum_{t=0}^{T_0-1} \|Err_t\|_2^2 + \|\mathcal{H}(x_t) - y_t\|_2^2 \ . \tag{C.4}$$

Choosing $T_0 = T$, taking expectation with respect to $\{(x_t, y_t)\}_{t=0,1,\ldots,T-1}$ on both sides, and using Claim C.6 (by noticing $O(\tau_v + \alpha \mathfrak{L}_{\mathcal{G}}) \leq 0.1$) and the definition of $\mathsf{OPT}$, we have

$$\frac{1}{4T} \sum_{t=0}^{T-1} \mathbb{E}_{(x,y) \sim \mathcal{D}} \|\mathcal{H}(x) - \mathsf{out}(\mathbf{W}_t, \mathbf{V}_t; x)\|_2^2 \leq \frac{\|\mathbf{W}^*\|_F^2}{2\eta_w T} + \frac{\|\mathbf{V}^*\|_F^2}{2\eta_v T} + O(\mathsf{OPT} + \delta_0)$$

where

$$\delta_0 = \Theta\left(\widetilde{\alpha}^2 + \tau_v^2(1 + \mathfrak{B}_{\mathcal{F}}) + \alpha \tau_v \mathfrak{L}_{\mathcal{G}}(\mathfrak{B}_{\mathcal{F}} + 1)\right)^2$$
$$= \widetilde{\Theta}\left(\widetilde{\alpha}^4 + \alpha^4 (kp_{\mathcal{G}} \mathfrak{C}_{\mathfrak{s}}(\mathcal{G}))^4 (1 + \mathfrak{B}_{\mathcal{F}})^2 + \alpha^4 (kp_{\mathcal{G}} \mathfrak{C}_{\mathfrak{s}}(\mathcal{G}))^2 \mathfrak{L}_{\mathcal{G}}^2 (\mathfrak{B}_{\mathcal{F}} + 1)^2\right)$$
$$= \widetilde{\Theta}\left(\alpha^4 (kp_{\mathcal{G}} \mathfrak{C}_{\mathfrak{s}}(\mathcal{G}))^4 (1 + \mathfrak{B}_{\mathcal{F}})^2\right)$$

Above, the last inequality uses $\frac{1}{kp_{\mathcal{G}} \mathfrak{C}_{\mathfrak{s}}(\mathcal{G})} \leq O(\frac{1}{1+\mathfrak{L}_{\mathcal{G}}})$ (see Fact A.3) and the choice of $\widetilde{\alpha}$ from Lemma C.2.

Using $\|\mathbf{W}^*\|_F \leq \tau_w/10$, $\|\mathbf{V}^*\|_F \leq \tau_v/10$, we have as long as $\delta \geq \mathsf{OPT} + \delta_0$,

$$\frac{1}{T} \sum_{t=0}^{T-1} \mathbb{E}_{(x,y) \sim \mathcal{D}} \|\mathcal{H}(x) - \mathsf{out}(\mathbf{W}_t, \mathbf{V}_t; x)\|_2^2 \leq O(\delta) \text{ as long as } T \geq \Omega(\frac{\tau_w^2/\eta_w + \tau_v^2/\eta_v}{\delta}).$$

Finally, we need to check that (C.3) holds. To do so, we use $\|Err_t\|_2 \leq O(\tau)$ from Claim C.6 and apply martingale concentration on (C.4) and derive that, with high probability

$$\frac{\|\mathbf{W}_{T_0} - \mathbf{W}^*\|_F^2}{2\eta_w T_0} + \frac{\|\mathbf{W}_{T_0} - \mathbf{V}^*\|_F^2}{2\eta_v T_0} \leq \frac{\|\mathbf{W}^*\|_F^2}{2\eta_w T_0} + \frac{\|\mathbf{V}^*\|_F^2}{2\eta_v T_0} + O(\delta) + \widetilde{O}\left(\frac{\tau_w}{\sqrt{T_0}}\right) \ .$$

This implies

$$\frac{\|\mathbf{W}_{T_0}\|_F^2}{4\eta_w T_0} + \frac{\|\mathbf{W}_{T_0}\|_F^2}{4\eta_v T_0} \leq \frac{\|\mathbf{W}^*\|_F^2}{\eta_w T_0} + \frac{\|\mathbf{V}^*\|_F^2}{\eta_v T_0} + O(\delta) + \widetilde{O}\left(\frac{\tau_w}{\sqrt{T_0}}\right) \ .$$

Using $\|\mathbf{W}^*\|_F \leq \tau_w/10$ and $\|\mathbf{V}^*\|_F \leq \tau_v/10$, and using the relationship $\frac{\tau_w^2}{\eta_w} = \frac{\tau_v^2}{\eta_v}$, we have

$$\frac{\|\mathbf{W}_{T_0}\|_F^2}{\tau_w^2} + \frac{\|\mathbf{W}_{T_0}\|_F^2}{\tau_v^2} \leq \frac{4\|\mathbf{W}^*\|_F^2}{\tau_w^2} + \frac{4\|\mathbf{V}^*\|_F^2}{\tau_v^2} + 0.1 + \widetilde{O}\left(\frac{\eta_w \sqrt{T_0}}{\tau_w}\right) \ .$$

Therefore, choosing

$$T = \widetilde{\Theta}\left(\frac{\tau_w^2}{\min\{1, \delta^2\}}\right) \quad \eta_w = \widetilde{\Theta}\left(\min\{1, \delta\}\right) \leq 0.1$$

we can ensure that $\frac{\|\mathbf{W}_{T_0}\|_F^2}{\tau_w^2} + \frac{\|\mathbf{W}_{T_0}\|_F^2}{\tau_v^2} \leq 1$ with high probability for all $T_0 = 0, 1, \ldots, T-1$ (so (C.3) holds).

Finally, we note that it satisfies $\mathsf{poly}(\mathfrak{C}_{\widetilde{\alpha}}(\mathcal{F}), \mathfrak{C}_{\widetilde{\alpha}}(\mathcal{G}), \widetilde{\alpha}^{-1}) \leq \mathsf{poly}(\mathfrak{C}_\alpha(\mathcal{F}), \mathfrak{C}_\alpha(\mathcal{G}), p_\mathcal{F}, \alpha^{-1})$ with the choice $\widetilde{\alpha} = \frac{\alpha}{k(p_\mathcal{F}\mathfrak{C}_s(\mathcal{F}) + p_\mathcal{G}\mathfrak{C}_s(\mathcal{G}))}$. $\square$

## D  Theorem 2 and Theorem 3 Proof Details

Our proof relies on the following two structural lemmas. The first one is a simple corollary of the Parseval's equality from Boolean analysis.

**Lemma D.1.** *For every $k \in \{2, 3, \cdots, d\}$, for every function $f(x) = \sum_{\mathcal{S}' \subseteq [d]} \lambda_{\mathcal{S}'} \prod_{j \in \mathcal{S}'} x_j$, suppose there exists $\mathcal{S} \subseteq [d]$ of size $k$ and $i \in \mathcal{S}$ such that*

$$\mathbb{E}_{x \sim U(\{-1,1\}^d)} \left[|f(x) - (x_i + \alpha \prod_{j \in \mathcal{S}} x_j)|^2\right] \leq \tfrac{1}{16}\alpha^2 \tag{D.1}$$

*Then we must have $\lambda_\mathcal{S} \geq \tfrac{3}{4}\alpha$ and $\sum_{\mathcal{S}' \subseteq [d], |\mathcal{S}'|=k, \mathcal{S}' \neq \mathcal{S}} \lambda_{\mathcal{S}'}^2 \leq \tfrac{1}{16}\alpha^2$.*

*Proof of Lemma D.1.* The lemma follows from the following equality that can be easily verified:

$$\mathbb{E}_{x \sim U(\{-1,1\}^d)}\left[|f(x) - (x_i + \alpha \prod_{j \in \mathcal{S}} x_j)|^2\right] = (\lambda_{\{i\}} - 1)^2 + (\lambda_\mathcal{S} - \alpha)^2 + \sum_{\mathcal{S}' \subseteq [d], \mathcal{S}' \neq \mathcal{S}, \mathcal{S}' \neq \{i\}} \lambda_{\mathcal{S}'}^2 \ . \quad \square$$

The next one can be proved by carefully bounding the matrix rank (see Section D.3).

**Lemma D.2.** *For every $\alpha > 0$, for every matrix $\mathbf{M} \in \mathbb{R}^{N \times R}$ where $R \geq 2N$, then there can not be vectors $a_1, \cdots, a_R \in \mathbb{R}^N$ such that for every $r \in [R]$:*

$$\langle \mathbf{M}_r, a_r \rangle \geq \tfrac{3}{4}\alpha \quad \text{and} \quad \sum_{r' \in [R], r' \neq r} \langle \mathbf{M}_{r'}, a_r \rangle^2 \leq \tfrac{1}{16}\alpha^2 \ .$$

### D.1  Proof of Theorem 2

Throughout the proof of Theorem 2, for notational simplicity, we re-scale inputs $x$ by $\sqrt{d}$ so that $x \in \{\pm 1\}^d$, and also re-scale $\mathbf{W}^*$ in the target function (7.1) to $\mathbf{W}^* = (\mathbf{e}_{i_1}, \mathbf{e}_{i_2}, \cdots \mathbf{e}_{i_k})$.

For notation simplicity, below we restate Theorem 2 with respect to one single output $k = 1$ and $d_1 = d$. The full statement for multiple outputs and more general distributions is a simple corollary (see Remark D.3).

**Theorem 2** (simplified). *For every integers $k, d, N$ satisfying $2 \leq k \leq d$ and $N \leq \frac{1}{1000}\binom{d_1}{k}$, for every (Mercer) kernel $K(x,y) \colon \mathbb{R}^{d \times d} \to \mathbb{R}$, for every $x^{(1)}, \cdots, x^{(N)} \in \mathbb{R}^d$, there exist at least $0.99 \times \binom{d}{k}$ many $\mathcal{S} \subseteq [d]$ of size $k$ such that, for every $i \in \mathcal{S}$, for every $w \in \mathbb{R}^N$ and the associated kernel function $\mathfrak{K}(x) = \sum_{n \in [N]} K(x, x^{(n)}) w_n$,*

$$\mathbb{E}_{x \sim U(\{-1,1\}^d)} \left[ \left| \mathfrak{K}(x) - \left( x_i + \alpha \prod_{j \in \mathcal{S}} x_j \right) \right|^2 \right] > \tfrac{1}{16} \alpha^2 \ .$$

*Proof of Theorem 2.* By property of (mercer) kernel, there exists feature mapping $\Phi(x) = (\phi_\ell(x))_{\ell \in \mathbb{N}}$ where each $\phi_\ell \colon \mathbb{R}^d \to \mathbb{R}$ such that:

$$K(x,y) = \sum_{\ell \in \mathbb{N}} \phi_\ell(x) \phi_\ell(y) \ .$$

Since we only care $x \in \{-1,1\}^d$, we can write each $\phi_\ell(x)$ in its (Boolean) Fourier basis:

$$\forall x \in \{-1,1\}^d \colon \quad \phi_\ell(x) = \sum_{\mathcal{S} \subseteq [d]} \lambda_{\mathcal{S},\ell} \prod_{j \in \mathcal{S}} x_j \ .$$

Given arbitrary $x^{(1)}, \ldots, x^{(N)} \in \mathbb{R}^d$, we can define matrix $\mathbf{M} \in \mathbb{R}^{N \times \binom{d}{k}}$ as follows:

$$\forall n \in [N], \forall \mathcal{S} \subseteq [d] \text{ with } |\mathcal{S}| = k \colon \quad \mathbf{M}_{n,\mathcal{S}} \overset{\text{def}}{=} \sum_{\ell \in \mathbb{N}} \lambda_{\mathcal{S},\ell} \phi_\ell(x^{(n)}) \ .$$

For any $w \in \mathbb{R}^N$, we can write

$$\mathfrak{K}(x) = \sum_{n \in [N]} K(x, x^{(n)}) w_n = \sum_{n \in [N]} \sum_{\ell \in \mathbb{N}} \phi_\ell(x) \phi_\ell(x^{(n)}) w_n$$
$$= \sum_{\mathcal{S}' \subseteq [d]} \left( \sum_{n \in [N]} \sum_{\ell \in \mathbb{N}} \lambda_{\mathcal{S}',\ell} \phi_\ell(x^{(n)}) w_n \right) \prod_{j \in \mathcal{S}'} x_j = \sum_{\mathcal{S}' \subseteq [d]} \langle \mathbf{M}_{\mathcal{S}'}, w \rangle \cdot \prod_{j \in \mathcal{S}'} x_j \quad (\text{D.2})$$

Hence, by Lemma D.1, if for some $\mathcal{S} \subseteq [d]$ of size $k$, there exists $i \in \mathcal{S}$ and exists $w_\mathcal{S} \in \mathbb{R}^N$ with $\mathfrak{K}_\mathcal{S}(x) = \sum_{n \in [N]} K(x, x^{(n)}) [w_\mathcal{S}]_n$ satisfying

$$\mathbb{E}_{x \sim U(\{-1,1\}^d)} \left[ |\mathfrak{K}_\mathcal{S}(x) - (x_i + \alpha \prod_{j \in \mathcal{S}} x_j)|^2 \right] \leq \frac{1}{16} \alpha^2 \ ,$$

then it must satisfy

$$\langle \mathbf{M}_\mathcal{S}, w_\mathcal{S} \rangle \geq \tfrac{3}{4} \alpha \quad \text{and} \quad \sum_{|\mathcal{S}| \subseteq [d], |\mathcal{S}'| = k, \mathcal{S}' \neq \mathcal{S}} \langle \mathbf{M}_{\mathcal{S}'}, w_\mathcal{S} \rangle^2 \leq \tfrac{1}{16} \alpha^2 \ .$$

However, according to Lemma D.2, as long as $\binom{d}{k} \geq 1000N$, we know that the above condition cannot hold for at least 0.99 fraction of the $\mathcal{S} \subseteq [d]$ of size $k$. This completes the proof. □

*Remark* D.3. In the full statement of Theorem 2, there are multiple outputs $\mathfrak{K}_1(x), \ldots, \mathfrak{K}_k(x)$. It suffices to focus on an arbitrary (say the first) coordinate and then apply the above lower bound.

In the full statement of Theorem 2, we have $x \sim \mathcal{D} \overset{\text{def}}{=} U(\{-1,1\}^{d_1}) \times \mathcal{D}_2$ for $k \leq d_1 \leq d$. In such a case, one can write each $x = (x_\triangleleft, x_\triangleright)$ for $x_\triangleleft \in \mathbb{R}^{d_1}$ and $x_\triangleright \in \mathbb{R}^{d-d_1}$. Now, equation (D.2) becomes

$$\mathbb{E}_{x_\triangleright \sim \mathcal{D}_2}[\mathfrak{K}(x)] = \mathbb{E}_{x_\triangleright \sim \mathcal{D}_2}[\mathfrak{K}(x_\triangleleft, x_\triangleright)] = \sum_{\mathcal{S}' \subseteq [d_1]} \langle \mathbf{M}_\mathcal{S}, w \rangle \cdot \prod_{j \in \mathcal{S}'} x_j$$

and the final statement can be derived using the following simple property, for every $\mathcal{S} \subseteq [d_1]$

$$\mathbb{E}_{(x_\triangleleft, x_\triangleright) \sim \mathcal{D}} \left[ |\mathfrak{K}(x_\triangleleft, x_\triangleright) - (x_i + \alpha \prod_{j \in \mathcal{S}} x_j)|^2 \right] \geq \mathbb{E}_{x_\triangleleft \sim U(\{-1,1\}^{d_1})} \left[ | \mathbb{E}_{x_\triangleright \sim \mathcal{D}_2}[\mathfrak{K}(x_\triangleleft, x_\triangleright)] - (x_i + \alpha \prod_{j \in \mathcal{S}} x_j)|^2 \right] \ .$$

## D.2 Proof of Theorem 3

For notation simplicity, we re-scale inputs $x$ by $\sqrt{d}$ so that $x \in \{\pm 1\}^d$, and also re-scale $\mathbf{W}^*$ in the target function (7.1) to $\mathbf{W}^* = (\mathbf{e}_{i_1}, \mathbf{e}_{i_2}, \cdots \mathbf{e}_{i_k})$.

Again for notation simplicity, below we restate Theorem 3 with respect to one single output $k = 1$ and $d_1 = d$. The full statement for multiple outputs and more general distributions is analogous (in the same spirit as Remark D.3).

**Theorem 3** (simplified). *For every integers $k, d, D$ satisfying $2 \leq k \leq d$ and $D \leq \frac{1}{1000}\binom{d}{k}$, for every $\alpha \in (0,1)$, for every feature mapping $\phi \colon \mathbb{R}^d \to \mathbb{R}^D$, there exist at least $0.99 \times \binom{d}{k}$ many $\mathcal{S} \subseteq [d]$ of size $k$ such that, for every $i \in \mathcal{S}$, for every $w \in \mathbb{R}^D$ and the associated linear function $\mathfrak{F}(x) = w^\top \phi(x)$,*

$$\mathbb{E}_{x \sim U(\{-1,1\}^d)}\left[\left|\mathfrak{F}(x) - \left(x_i + \alpha \prod_{j \in \mathcal{S}} x_j\right)\right|^2\right] > \tfrac{1}{16}\alpha^2 \ .$$

*Proof of Theorem 3.* Let us write $\phi(x) = (\phi_1(x), \cdots, \phi_D(x))$ where each $\phi_i \colon \mathbb{R}^d \to \mathbb{R}$. Since we only focus on $x \in \{-1, 1\}^d$ we can write

$$\phi_i(x) = \sum_{\mathcal{S} \subseteq [d]} \lambda_{\mathcal{S},i} \prod_{j \in \mathcal{S}} x_j$$

for some set of coefficients $\lambda_{\mathcal{S},i} \in \mathbb{R}$. Now, define matrix $\mathbf{M} \in \mathbb{R}^{D \times 2^d}$ as follows:

$$\forall i \in [D], \quad \forall \mathcal{S} \subseteq [d] \colon \quad \mathbf{M}_{i,\mathcal{S}} = \lambda_{\mathcal{S},i} \ .$$

We have for every $w \in \mathbb{R}^D$ (that can possibly depend on $\mathcal{S}$),

$$w^\top \phi(x) = \sum_{\mathcal{S} \subseteq [d]} \langle \mathbf{M}_\mathcal{S}, w \rangle \prod_{j \in \mathcal{S}} x_j$$

This is exactly (D.2) in the proof of Theorem 2, so the rest of the proof follows analogously by applying Lemma D.2. □

## D.3 Proof of Lemma D.2

*Proof of Lemma D.2.* Suppose by way towards contradiction that there exist vectors $a_1, \cdots, a_R \in \mathbb{R}^N$ such that for every $r \in [R]$:

$$\langle \mathbf{M}_r, a_r \rangle \geq \tfrac{3}{4}\alpha \quad \text{and} \quad \sum_{r' \in [R], r' \neq r} \langle \mathbf{M}_{r'}, a_r \rangle^2 \leq \tfrac{1}{16}\alpha^2$$

Let us define $b_r = \frac{1}{\langle \mathbf{M}_r, a_r \rangle} a_r$ so they become

$$\langle \mathbf{M}_r, b_r \rangle = 1 \quad \text{and} \quad \sum_{r' \in [R], r' \neq r} \langle \mathbf{M}_{r'}, b_r \rangle^2 \leq \tfrac{1}{9}$$

Now, defining matrix $\mathbf{B} = \{b_r\}_{r \in [R]} \in \mathbb{R}^{N \times R}$, we can rewrite

$$\mathbf{B}^\top \mathbf{M} = \mathbf{I} + \mathbf{E} \in \mathbb{R}^{R \times R} \tag{D.3}$$

where $\mathbf{E}$ is matrix with zero diagonals. Since for every $r \in [R]$, it satisfies $\sum_{r' \in [R]} \mathbf{E}_{r,r'}^2 = \sum_{r' \in [R], r' \neq r} \langle \mathbf{M}_{r'}, b_r \rangle^2 \leq \tfrac{1}{9}$, we conclude that $\|\mathbf{E}\|_F^2 \leq \tfrac{1}{9}R$.

Next, since $\mathbf{E}$ cannot have more than $\tfrac{1}{9}R$ singular values that are $\geq 1$. By the min-max theorem for singular values (a.k.a. Courant-Fischer theorem), there exists a subspace $U$ of $\mathbb{R}^R$ with dimension $\tfrac{8}{9}R$ such that $\max_{x \in U, \|x\|_2 = 1} \|\mathbf{E}x\|_2 < 1$. As a result, for every non-zero $x \in U$, we have $\|(\mathbf{I} + \mathbf{E})x\|_2 \geq \|x\|_2 - \|\mathbf{E}x\|_2 > 0$. This implies

$$\operatorname{rank}(\mathbf{I} + \mathbf{E}) \geq \frac{8}{9}R \ .$$

To the contrary, we have $\operatorname{rank}(\mathbf{B}^\top \mathbf{M}) \leq N \leq \tfrac{1}{2}R$. This gives a contradiction. □

## D.4 Proof of Corollary 7.1

*Proof of Corollary 7.1.* To apply Theorem 1, we need to carefully verify Concept 1 by appropriately re-scaling. Without loss of generality suppose $(i_1, \ldots, i_k) = (1, \ldots, k)$. For every $i \in [k]$, let us define
$$z_i \stackrel{\text{def}}{=} \mathcal{F}_i(x) \stackrel{\text{def}}{=} \frac{\sqrt{d}}{\sqrt{k}} x_i$$
which satisfies $\mathfrak{C}_\mathfrak{s}(\mathcal{F}) = O(\sqrt{d})$, $p_\mathcal{F} = 1$, and $\|\mathcal{F}(x)\|_2 = 1$. Next, let us define
$$\mathcal{G}_r(z) \stackrel{\text{def}}{=} \frac{k^k}{\sqrt{k} k! 2^k} \sum_{s \in \{0,1\}^k} (-1)^{s_1 + \cdots + s_k} \Big( \sum_{i \in [k]} \frac{(-1)^{s_i} z_i}{\sqrt{k}} \Big)^k$$
and one can verify that $\mathcal{G}_r(z) = \frac{k^{k/2}}{\sqrt{k}} \prod_{i \in [k]} z_i$ and therefore $\mathcal{G}_r(\mathcal{F}(x)) = \frac{1}{\sqrt{k}} \prod_{i \in [k]} (\sqrt{d} x_i)$. It also satisfies $\mathfrak{C}_\mathfrak{s}(\mathcal{G}) = 2^{O(k)}$ and $p_\mathcal{G} = 2^k$. In sum, we have constructed
$$\mathcal{F}(x) + \alpha \mathcal{G}(\mathcal{F}(x)) = \tfrac{1}{\sqrt{k}} \Big( \sqrt{d} x_i + \alpha \prod_{j \in [k]} (\sqrt{d} x_j) \Big)_{i \in [k]}$$
and we can thus apply Theorem 1 (after rescaling the label by $1/\sqrt{k}$). $\square$

# E  Existential Tool

In this section we include a simple variant of the existential lemma from [4]. We include the proofs only for completeness' sake.

Consider random function $G((x, 1); \mathbf{W}^*) = (G_1((x, 1); \mathbf{W}^*), \ldots, G_k((x, 1); \mathbf{W}^*))$ in which
$$G_r((x, 1); \mathbf{W}^*) \stackrel{\text{def}}{=} \sum_{i=1}^m a_{r,i} \cdot \langle w_i^*, (x, 1) \rangle \cdot \mathbb{1}_{\langle w_i^{(0)}, (x,1) \rangle \geq 0}$$
where $\mathbf{W}^* \in \mathbb{R}^{m \times (d+1)}$ is a given matrix, $\mathbf{W}^{(0)} \in \mathbb{R}^{m \times (d+1)}$ is a random matrix where each $w_i^{(0)}$ is i.i.d. from $\mathcal{N}(0, \frac{\mathbf{I}}{m})$, and each $a_{r,i}$ is i.i.d. from $\mathcal{N}(0, 1)$.

We have the following main lemma of this section:

**Lemma E.1.** *Given any $\mathcal{F} \colon \mathbb{R}^d \to \mathbb{R}^k$ with general complexity $(p, \mathfrak{C}_\mathfrak{s}(\mathcal{F}), \mathfrak{C}_\varepsilon(\mathcal{F}))$, for every $\varepsilon \in (0, \frac{1}{pk\mathfrak{C}_\mathfrak{s}(\mathcal{F})})$, there exists $M = \mathsf{poly}(\mathfrak{C}_\varepsilon(\mathcal{F}), 1/\varepsilon)$ such that if $m \geq M$, then with high probability there is a construction $\mathbf{W}^* = (w_1^*, \ldots, w_m^*) \in \mathbb{R}^{m \times d}$ (that does not depend on x) with*
$$\|\mathbf{W}^*\|_{2,\infty} \leq \tfrac{kp\mathfrak{C}_\varepsilon(\mathcal{F})}{m} \quad \text{and} \quad \|\mathbf{W}^*\|_F \leq \widetilde{O}(\tfrac{kp\mathfrak{C}_\mathfrak{s}(\mathcal{F})}{\sqrt{m}})$$
*satisfying, for every $x \in \mathbb{R}^d$, with probability at least $1 - e^{-\Omega(\sqrt{m})}$*
$$\sum_{r=1}^k |\mathcal{F}_r(x) - G_r((x, 1); \mathbf{W}^*)| \leq \varepsilon \cdot \|(x, 1)\|_2,$$

## E.1  Restate Lemma E.1

We first note that, by replacing $(x, 1)$ with $x$, we can restate Lemma E.1 as follows. Consider a target function $\Phi \colon \mathbb{R}^d \to \mathbb{R}^k$ where
$$\Phi_r(x) \stackrel{\text{def}}{=} \sum_{i=1}^p a_{r,i}^* \cdot \phi_{r,i} \Big( \frac{\langle w_{1,r,i}^*, x \rangle}{\|x\|_2} \Big) \cdot \langle w_{2,r,i}^*, x \rangle$$

and $\phi_{r,i} \colon \mathbb{R} \to \mathbb{R}$ has only zero-order and odd-order terms in its Taylor expansion at zero, and $|a^*_{r,i}| \leq 1$, $\|w^*_{1,i}\|_2 = \|w^*_{2,i}\|_2 = 1$, $\mathfrak{C}_\varepsilon(\Phi) = \max_{r,i}\{\mathfrak{C}_\varepsilon(\phi_{r,i})\}$ and $\mathfrak{C}_\mathfrak{s}(\Phi) = \max_{r,i}\{\mathfrak{C}_\mathfrak{s}(\phi_{r,i})\}$. Let

$$G_r(x; \mathbf{W}^\circledast) \stackrel{\text{def}}{=} \sum_{i=1}^m a_{r,i} \cdot \langle w_i^\circledast, x \rangle \cdot \mathbb{1}_{\langle w_i^{(0)}, x \rangle \geq 0} \ .$$

be the similarly defined random function. We have the following:

**Lemma E.1'.** *For every $\varepsilon \in (0, \frac{1}{pk\mathfrak{C}_\mathfrak{s}(\Phi,1)})$, there exists $M = \mathsf{poly}(\mathfrak{C}_\varepsilon(\Phi,1), 1/\varepsilon)$ such that if $m \geq M$, then with high probability there is a construction $\mathbf{W}^\circledast = (w_1^\circledast, \ldots, w_m^\circledast) \in \mathbb{R}^{m \times d}$ (that does not depend on $x$) with*

$$\|\mathbf{W}^\circledast\|_{2,\infty} \leq \frac{kp\mathfrak{C}_\varepsilon(\Phi,1)}{m} \quad \text{and} \quad \|\mathbf{W}^\circledast\|_F \leq \widetilde{O}(\frac{kp\mathfrak{C}_\mathfrak{s}(\Phi,1)}{\sqrt{m}})$$

*satisfying, for every $x \in \mathbb{R}^d$, with probability at least $1 - e^{-\Omega(\sqrt{m})}$*

$$\sum_{r=1}^k \left| \Phi_r(x) - G_r(x; \mathbf{W}^\circledast) \right| \leq \varepsilon \cdot \|x\|_2,$$

We stress that Lemma E.1' is a modified version of Lemma G.1 from [4, ver.4]. The only difference is that in their original Lemma G.1, the indicator function $\mathbb{1}_{\langle w_i^{(0)}, x \rangle \geq 0}$ has an additional random bias term (that is, becomes $\mathbb{1}_{\langle w_i^{(0)}, x \rangle + b_i^{(0)} \geq 0}$). In our Lemma E.1', we do not allow such bias and thus we can only fit functions $\Phi$ whose Taylor expansions have only zero-order and odd-order terms (as opposed to arbitrary smooth functions in the original Lemma G.1).

The proof of Lemma E.1' is based on the following "indicator to function" lemma, which is a simple modification from Lemma 5.2 of [4, ver.4]. It says that given unit vector $w^* \in \mathbb{R}^d$, we can approximate function $\phi(\langle w^*, x \rangle)$ (over $x$) by designing a random function $\mathbb{1}_{\langle w, x \rangle \geq 0} h(\langle w, w^* \rangle)$ where $w$ is a random Gaussian and $h(\cdot)$ is a function at our choice. Again, the only difference between our Lemma E.2 and Lemma 5.2 of [4, ver.4] is that we do not have the random bias term.

**Lemma E.2** (indicator to function). *For every smooth function $\phi$ that only has zero-order and odd-order terms in its Taylor expansion at point zero, every $\varepsilon \in \left(0, \frac{1}{\mathfrak{C}_\mathfrak{s}(\phi,1)}\right)$, there exists a function $h \colon \mathbb{R} \to [-\mathfrak{C}_\varepsilon(\phi,1), \mathfrak{C}_\varepsilon(\phi,1)]$ that is also $\mathfrak{C}_\varepsilon(\phi,1)$-Lipschitz continuous with the following two (equivalent) properties:*

*(a) For every $x_1 \in [-1,1]$:*

$$\left| \mathbb{E}\left[ \mathbb{1}_{\alpha_1 x_1 + \beta_1 \sqrt{1-x_1^2} \geq 0} h(\alpha_1) \right] - \phi(x_1) \right| \leq \varepsilon$$

*where $\alpha_1, \beta_1 \sim \mathcal{N}(0,1)$ are independent random variables.*

*(b) For every $w^*, x \in \mathbb{R}^d$ with $\|w^*\|_2 = \|x\|_2 = 1$:*

$$\left| \mathbb{E}\left[ \mathbb{1}_{\langle w, x \rangle \geq 0} h(\langle w, w^* \rangle) \right] - \phi(\langle w^*, x \rangle) \right| \leq \varepsilon$$

*where $w \sim \mathcal{N}(0, \mathbf{I})$ is an $d$-dimensional Gaussian.*

*Furthermore, $h$ satisfies $\mathbb{E}_{\alpha_1 \sim \mathcal{N}(0,1)}\left[ h(\alpha_1)^2 \right] \leq \mathfrak{C}_\mathfrak{s}(\phi,1)^2$.*

In the remainder of this section, for sake of completeness, we first prove Lemma E.2 in Section E.2, and then prove Lemma E.1' and Section E.3.

## E.2 Proof of Lemma E.2: Indicator to Function

Recall from [4] by renaming variables it suffices to prove Lemma E.2a. For notation simplicity, let us denote $w_0 = (\alpha_1, \beta_1)$ and $x = (x_1, \sqrt{1-x_1^2})$ where $\alpha_1, \beta_1$ are two independent random standard Gaussians.

Throughout the proof, we also take an alternative view of the randomness. We write $\langle w_0, x \rangle = \alpha$ and $\alpha_1 = \alpha x_1 + \sqrt{1-x_1^2}\beta$ for two independent $\alpha, \beta \sim \mathcal{N}(0,1)$.[20]

We first make a technical claim involving in fitting monomials in $x_1$. It is a simplified version of Claim B.1 of [4, ver.4].

**Claim E.3.** *Let $h_i(x)$ be the degree-$i$ Hermite polynomial (see Definition A.4 of [4, ver.4]). For every odd integer $i \geq 1$ there exists constant $p'_i$ with $|p'_i| \geq \frac{(i-1)!!}{4}$ such that*

$$x_1^i = \frac{1}{p'_i} \mathop{\mathbb{E}}_{w_0 \sim \mathcal{N}(0,\mathbf{I}) \sim \mathcal{N}(0,1)} [h_i(\alpha_1) \cdot \mathbb{1}[\langle x, w_0 \rangle \geq 0]]$$

(The proof of Claim E.3 is identical to that of the original Claim B.1 of [4, ver.4] by forcing the bias term $b_0 = 0$.)

We next use Claim E.3 to fit arbitrary functions $\phi(x_1)$. By Taylor expansion, we have

$$\phi(x_1) = c_0 + \sum_{i=1,\text{ odd } i}^{\infty} c_i x_1^i = c_0 + \sum_{i=1}^{\infty} c'_i \cdot \mathop{\mathbb{E}}_{\alpha, \beta, b_0 \sim \mathcal{N}(0,1)} [h_i(\alpha_1) \cdot \mathbb{1}[\langle x, w_0 \rangle + b_0 \geq 0]]$$

where

$$c'_i \stackrel{\text{def}}{=} \frac{c_i}{p'_i}, \quad |c'_i| \leq \frac{4|c_i|}{(i-1)!!} \tag{E.1}$$

Next, recall the following claim on absolute values of the Hermite polynomials (see Claim B.2 of [4, ver.4]).

**Claim E.4.** *Setting $B_i \stackrel{\text{def}}{=} 100 i^{1/2} + 10\sqrt{\log \frac{1}{\varepsilon}}$, we have*

*(a)* $\sum_{i=1}^{\infty} |c'_i| \cdot \mathbb{E}_{z \sim \mathcal{N}(0,1)} [|h_i(z)| \cdot \mathbb{1}[|z| \geq B_i]] \leq \epsilon/8$

*(b)* $\sum_{i=1}^{\infty} |c'_i| \cdot \mathbb{E}_{z \sim \mathcal{N}(0,1)} [|h_i(B_i)| \cdot \mathbb{1}[|z| \geq B_i]] \leq \epsilon/8$

*(c)* $\sum_{i=1}^{\infty} |c'_i| \cdot \mathbb{E}_{z \sim \mathcal{N}(0,1)} [|h_i(z)| \cdot \mathbb{1}[|z| \leq B_i]] \leq \frac{1}{2}\mathfrak{C}_\varepsilon(\phi, 1)$

*(d)* $\sum_{i=1}^{\infty} |c'_i| \cdot \mathbb{E}_{z \sim \mathcal{N}(0,1)} [|\frac{d}{dz} h_i(z)| \cdot \mathbb{1}[|z| \leq B_i]] \leq \frac{1}{2}\mathfrak{C}_\varepsilon(\phi, 1)$

Now, let us define $\widehat{h}_i(\alpha_1) \stackrel{\text{def}}{=} h_i(\alpha_1) \cdot \mathbb{1}[|\alpha_1| \leq B_i] + h_i(\text{sign}(\alpha_1) B_i) \cdot \mathbb{1}[|\alpha_1| > B_i]$ as the truncated version of the Hermite polynomial $h_i(\cdot)$. Using Claim E.4, we have

$$\phi(x_1) = c_0 + R'(x_1) + \sum_{i=1}^{\infty} c'_i \cdot \mathop{\mathbb{E}}_{\alpha, \beta \sim \mathcal{N}(0,1)} \left[\widehat{h}_i(\alpha_1) \cdot \mathbb{1}[\langle x, w_0 \rangle \geq 0]\right]$$

where $|R'(x_1)| < \epsilon/4$ uses Claim E.4a and Claim E.4b. In other words, if we define

$$h(\alpha_1) \stackrel{\text{def}}{=} 2c_0 + \sum_{i=1}^{\infty} c'_i \cdot \widehat{h}_i(\alpha_1)$$

---

[20]This is possible for the following reason. Let $x^\perp = (\sqrt{1-x_1^2}, -x_1)$ be unit vector orthogonal to $x$. We can write $w_0 = \alpha x + \beta x^\perp$ where $\alpha, \beta \sim \mathcal{N}(0,1)$ are two independent Gaussians.

then we have
$$\left| \mathop{\mathbb{E}}_{\alpha,\beta \sim \mathcal{N}(0,1)} \left[ \mathbb{1}[\langle x, w_0 \rangle \geq 0] \cdot h(\alpha_1) \right] - \phi(x_1) \right| = |R'(x_1)| \leq \varepsilon/4 \ .$$

As for the range of $h$, we use Claim E.4b and Claim E.4c to derive that
$$|h(\alpha_1)| \leq 2c_0 + \frac{\varepsilon}{8} + \frac{1}{2}\mathfrak{C}_\varepsilon(\phi, 1) \leq \mathfrak{C}_\varepsilon(\phi, 1) \ .$$

As for the Lipschitz continuity of $h$ on its first coordinate $\alpha_1$, we observe that for each $i > 0$, $\widehat{h}_i(z)$ has zero sub-gradient for all $|z| \geq B_i$. Therefore, it suffices to bound $\left|\frac{d}{dz}h_i(z)\right|$ for $|z| < B_i$. Replacing the use of Claim E.4c by Claim E.4d immediately gives us the same bound on the Lipschitz continuity of $h$ with respect to $\alpha_1$.

As for the expected square $\mathbb{E}_{\alpha_1 \sim \mathcal{N}(0,1)}\left[h(\alpha_1)^2\right]$, we can write
$$h(\alpha_1) = 2c_0 + \sum_{i=1}^{\infty} c'_i \cdot \widehat{h}_i(\alpha_1) \stackrel{\text{①}}{=} 2c_0 + \sum_{i=1}^{\infty} c'_i \cdot h_i(\alpha_1) \pm \frac{\varepsilon}{4}$$

Above, ① uses Claim E.4a and Claim E.4b. Using the othogonality condition of Hermite polynomials (that is, $\mathbb{E}_{x \sim \mathcal{N}(0,1)}[h_i(x)h_j(x)] = \sqrt{2\pi} j! \delta_{i,j}$), we immediately have

$$\mathop{\mathbb{E}}_{\alpha_1 \sim \mathcal{N}(0,1)}[h(\alpha_1)^2] \leq O(\varepsilon^2 + c_0^2) + O(1) \cdot \sum_{i=1}^{\infty} (c'_i)^2 (i!)$$
$$\leq O(\varepsilon^2 + c_0^2) + O(1) \cdot \sum_{i=1}^{\infty} \frac{(i!) \cdot |c_i|^2}{((i-1)!!)^2}$$
$$\leq O(\varepsilon^2 + c_0^2) + O(1) \cdot \sum_{i=1}^{\infty} i^{0.5} \cdot |c_i|^2 \leq \mathfrak{C}_\mathfrak{s}(\phi, 1)^2 \ .$$

Above, ① uses inequality $\frac{i!}{((i-1)!!)^2} \leq 2\sqrt{i}$ for all $i \geq 1$.

This finishes the proof of Lemma E.2a. ∎

### E.3 Proof of Lemma E.1'

Without loss of generality we assume $\|x\|_2 = 1$ in this proof. (Both $\Phi$ and $G$ are positive homogeneous in $x$.)

**Fit a single function** $a^*_{r,i}\phi_{r,i}(\langle w^*_{1,r,i}, x\rangle)\langle w^*_{2,r,i}, x\rangle$. We first fix some $r \in [k]$ and $i \in [p]$ and construct weights $w^{\circledast}_j \in \mathbb{R}^d$. Let $h^{(r,i)}(\cdot)$ be the function $h(\cdot)$ constructed from $\phi = \phi_{r,i}$ using Lemma E.2. We have $|h^{(r,i)}| \leq \mathfrak{C}_\varepsilon(\Phi, 1)$. Define
$$w^{\circledast}_j \stackrel{\text{def}}{=} a_{r,j} a^*_{r,i} h^{(r,i)}\left(\sqrt{m}\langle w^{(0)}_j, w^*_{1,i}\rangle\right) w^*_{2,i} \tag{E.2}$$

where $\sqrt{m}\langle w^{(0)}_j, w^*_{1,i}\rangle$ has the same distribution with $\alpha_1$ in Lemma E.2. By Lemma E.2, we have that
$$\mathop{\mathbb{E}}_{w^{(0)}_j, a_{r,j}}\left[a_{r,j}\mathbb{1}_{\langle w^{(0)}_j, x\rangle \geq 0}\langle w^{\circledast}_j, x\rangle\right] = \mathop{\mathbb{E}}_{w^{(0)}_j}\left[a^*_{r,i}\mathbb{1}_{\langle w^{(0)}_j, x\rangle \geq 0} h^{(r,i)}\left(\sqrt{m}\langle w^{(0)}_j, w^*_{1,i}\rangle\right)\langle w^*_{2,i}, x\rangle\right]$$
$$= a^*_{r,i}\phi_{r,i}(\langle w^*_{1,i}, x\rangle)\langle w^*_{2,i}, x\rangle \pm \varepsilon \ .$$

**Fit a combination** $\sum_{i \in [p]} a^*_{r,i}\phi_{r,i}(\langle w^*_{1,r,i}, x\rangle)\langle w^*_{2,r,i}, x\rangle$. We can re-define (the norm grows by

a maximum factor of $p$)
$$w_j^\circledast = a_{r,j} \sum_{i \in [p]} a_{r,i}^* h^{(r,i)} \left(\sqrt{m}\langle w_j^{(0)}, w_{1,i}^*\rangle\right) w_{2,i}^*$$

and the same above argument gives
$$\mathop{\mathbb{E}}_{w_j^{(0)}, a_{r,j}} \left[a_{r,j} \mathbb{1}_{\langle w_j^{(0)}, x\rangle \geq 0} \langle w_j^\circledast, x\rangle\right] = \sum_{i \in [p]} a_{r,i}^* \phi_{r,i}(\langle w_{1,i}^*, x\rangle) \langle w_{2,i}^*, x\rangle \pm \varepsilon p.$$

**Fit multiple outputs.** If there are $k$ outputs let us re-define (the norm grows by a maximum factor of $k$)
$$w_j^\circledast = \sum_{r \in [k]} a_{r,j} \sum_{i \in [p]} a_{r,i}^* h^{(r,i)} \left(\sqrt{m}\langle w_j^{(0)}, w_{1,i}^*\rangle\right) w_{2,i}^*. \tag{E.3}$$

and consider the quantity
$$\Xi_{r,j} \stackrel{\text{def}}{=} a_{r,j} \langle w_j^\circledast, x\rangle \mathbb{1}_{\langle w_j^{(0)}, x\rangle \geq 0} \ .$$

By randomness of $a$ we know that for $r' \neq r$, $\mathbb{E}[a_{r,j} a_{r',j}] = 0$. Thus, for every $r \in [k]$, it satisfies

$$\mathop{\mathbb{E}}_{w_j^{(0)}, a_{1,j}, \dots, a_{k,j}} [\Xi_{r,j}]$$
$$= \mathop{\mathbb{E}}_{w_j^{(0)}, a_{1,j}, \dots, a_{k,j}} \left[\sum_{r' \in [k]} a_{r,j} a_{r',j} \sum_{i \in [p]} \mathbb{1}_{\langle w_j^{(0)}, x\rangle \geq 0} a_{r',i}^* h^{(r,i)} \left(\sqrt{m}\langle w_j^{(0)}, w_{1,i}^*\rangle\right) \langle w_{2,i}^*, x\rangle\right]$$
$$= \mathop{\mathbb{E}}_{w_j^{(0)}} \left[\sum_{i \in [p]} \mathbb{1}_{\langle w_j^{(0)}, x\rangle \geq 0} a_{r,i}^* h^{(r,i)} \left(\sqrt{m}\langle w_j^{(0)}, w_{1,i}^*\rangle\right) \langle w_{2,i}^*, x\rangle\right]$$
$$= \sum_{i \in [p]} a_{r,i}^* \phi_{r,i}(\langle w_{1,i}^*, x\rangle) \langle w_{2,i}^*, x\rangle \pm p\varepsilon = \Phi_r^*(x) \pm p\varepsilon \ .$$

Now, re-scaling each $w_j^\circledast$ by a factor of $\frac{1}{m}$ and re-scaling $\varepsilon$ by $\frac{1}{2pk}$, we can write
$$G_r(x; \mathbf{W}^\circledast) = \sum_{j=1}^m \Xi_{r,j} \quad \text{and} \quad \mathbb{E}\left[G_r(x; \mathbf{W}^\circledast)\right] = \Phi_r^*(x) \pm \frac{\varepsilon}{2k} \ .$$

Now, we use $|h^{(r,i)}| \leq \mathfrak{C}_\varepsilon(\Phi, 1)$ and apply the concentration from Lemma B.3, which implies for our parameter choice of $m$, with probability at least $1 - e^{-\Omega(m\varepsilon^2/(k^4 p^2 \mathfrak{C}_\varepsilon(\Phi, 1)))}$
$$|G_r(x; \mathbf{W}^\circledast) - \Phi_r^*(x)| \leq \frac{\varepsilon}{k} \ .$$

**Norm on $\mathbf{W}^\circledast$.** According to its definition in (E.3), we have for each $j \in [m]$, with high probability $\|w_j^\circledast\|_2 \leq \widetilde{O}\left(\frac{kp\mathfrak{C}_\varepsilon(\Phi,1)}{m}\right)$ (here the additional $\frac{1}{m}$ is because we have re-scaled $w_j^\circledast$ by $\frac{1}{m}$). This means $\|\mathbf{W}^\circledast\|_{2,\infty} \leq \widetilde{O}\left(\frac{kp\mathfrak{C}_\varepsilon(\Phi,1)}{m}\right)$. As for the Frobenius norm,
$$\|\mathbf{W}^\circledast\|_F^2 = \sum_{j \in [m]} \|w_j^\circledast\|_2^2 \leq \sum_{j \in [m]} \widetilde{O}\left(\frac{k^2 p}{m^2}\right) \cdot \sum_{i \in [p]} h^{(r,i)} \left(\sqrt{m}\langle w_j^{(0)}, w_{1,i}^*\rangle\right)^2 \tag{E.4}$$

Now, for each $i \in [p]$, we know that $\sum_{j \in [m]} h^{(r,i)} \left(\sqrt{m}\langle w_j^{(0)}, w_{1,i}^*\rangle\right)^2$ is a summation of i.i.d. random variables, each with expectation at most $\mathfrak{C}_\mathfrak{s}(\Phi, 1)^2$ by Lemma E.2. Applying Hoeffding's concen-tration, we have with probability at least $1 - e^{-\Omega(\sqrt{m})}$

$$\sum_{j \in [m]} h^{(r,i)} \left( \sqrt{m} \langle w_j^{(0)}, w_{1,i}^* \rangle, \sqrt{m} b_j^{(0)} \right)^2 \leq m \cdot \mathfrak{C}_{\mathfrak{s}}(\Phi, 1)^2 + m^{3/4} \cdot \mathfrak{C}_{\varepsilon}(\Phi, 1)^2 \leq 2m \mathfrak{C}_{\mathfrak{s}}(\Phi, 1)^2$$

Putting this back to (E.4) we have $\|\mathbf{W}^*\|_F^2 \leq \widetilde{O}(\frac{k^2 p^2 \mathfrak{C}_{\mathfrak{s}}(\Phi,1)^2}{m})$. This finishes the proof of Lemma E.1'.